\documentclass[a4paper,5pt,authoryear]{elsarticle}
\usepackage{booktabs}
\usepackage{multicol}
\usepackage{ragged2e}
\usepackage{graphicx}
\usepackage{subfigure}
\usepackage{geometry}
\usepackage{ragged2e}
\usepackage{multirow}
\usepackage{multicol}
\usepackage{svg}

\usepackage{booktabs}   
\usepackage{array}      
\usepackage{tabularx}  
\usepackage{float}      
\usepackage{amsmath}    
 \usepackage{makecell}
\usepackage{natbib}

\usepackage{algorithm}
\usepackage{algpseudocode}

\algrenewcommand\algorithmicrequire{\textbf{Input:}}
\algrenewcommand\algorithmicensure{\textbf{Output:}}
\algrenewcommand\algorithmiccomment[1]{\hfill$\triangleright$~#1}
\algrenewcommand\alglinenumber[1]{\footnotesize #1:\,}
\newcommand{\StepHeaderLen}[2]{
	\Statex \hspace*{#1}\textcolor{blue}{\texttt{/* #2 */}}
}

\usepackage[colorlinks=true,
            linkcolor=blue,
            anchorcolor=blue,
            citecolor=blue,
			CJKbookmarks=true
            ]{hyperref}
\usepackage{amssymb}
\usepackage{color}

\journal{Knowledge-Based Systems}

\begin{document}

\begin{frontmatter}


\small


\title{A Multi-Mode Structured Light 3D Imaging System with Multi-Source Information Fusion for Underwater Pipeline Detection}

%
\author[buct]{Qinghan Hu}
\author[buct]{Haijiang Zhu\corref{cor1}}
\ead{zhuhj@mail.buct.edu.cn}
\author[ts]{Na Sun}
\author[gn]{Lei Chen}
\author[bua]{Zhengqiang Fan}
\author[buct]{Zhiqing Li\corref{cor1}}
\ead{lizhiqing@buct.edu.cn}
\cortext[cor1]{Corresponding authors.}
\address[buct]{College of Information Science and Technology, Beijing University of Chemical Technology, Beijing 100029, China}
\address[ts]{Department of Mechanical Engineering, Tsinghua University, Beijing 100084, China }
\address[gn]{Guoneng Zhishen Control Technology Co., Ltd., Beijing 102211, China }
\address[bua]{College of Intelligent Science and Engineering, Beijing University of Agriculture, Beijing 102206, China}

\begin{abstract}
\indent Underwater pipelines are highly susceptible to corrosion, which not only shorten their service life but also pose significant safety risks. Compared with manual inspection, the intelligent real-time imaging system for underwater pipeline detection has become a more reliable and practical solution. Among various underwater imaging techniques, structured light 3D imaging can restore the sufficient spatial detail for precise defect characterization. Therefore, this paper develops a multi-mode underwater structured light 3D imaging system for pipeline detection (UW-SLD system) based on multi-source information fusion. First, a rapid distortion correction (FDC) method is employed for efficient underwater image rectification. To overcome the challenges of extrinsic calibration among underwater sensors, a factor graph-based parameter optimization method is proposed to estimate the transformation matrix between the structured light and acoustic sensors. Furthermore, a multi-mode 3D imaging strategy is introduced to adapt to the geometric variability of underwater pipelines. Given the presence of numerous disturbances in underwater environments, a multi-source information fusion strategy and an adaptive extended Kalman filter (AEKF) are designed to ensure stable pose estimation and high-accuracy measurements.  In particular, an edge detection-based ICP (ED-ICP) algorithm is proposed. This algorithm integrates pipeline edge detection network with enhanced point cloud registration to achieve robust and high-fidelity reconstruction of defect structures even under variable motion conditions. Extensive experiments are conducted under different operation modes, velocities, and depths. The results demonstrate that the developed system achieves superior accuracy, adaptability and robustness, providing a solid foundation for autonomous underwater pipeline detection.
\end{abstract}


\end{frontmatter}

\begin{multicols}{2}
{
\section{Introduction}
\label{section：1}
\indent Underwater pipelines serve as critical carriers for fluid transportation in various marine engineering applications, including oil and gas delivery, seawater desalination, and wastewater reutilization projects~\citep{zhang2024exploring1}.  In addition, they play an indispensable role in underwater agricultural systems such as deep-sea aquaculture~\citep{li2025technological2}. However, prolonged exposure to complex marine environments subjects pipelines to various environmental stresses, including waves, tides, currents, and corrosion~\citep{xia2025deterioration3}. These factors can gradually lead to structural degradation and significantly compromise operational safety. Without effective inspection and maintenance, such degradation can lead to catastrophic failures, such as leakage or explosion, causing serious economic and environmental consequences~\citep{sharma2024comprehensive4}. Among various environmental factors, corrosion is widely recognized as the primary cause of structural degradation~\citep{hussein2023analysis5}. Previous studies revealed that corrosion markedly reduces the stable load-bearing capacity of underwater steel structures.  In their findings, the stable load-bearing capacity of corroded areas were approximately 24.7\% and 37.3\% lower than those of non-corroded regions after 5 and 10 years of service~\citep{xia2023non6}. This demonstrates the pronounced long-term impact of corrosion on structural integrity. Therefore, timely detection and maintenance of underwater pipeline corrosion are vital to safe operation and long-term reliability~\citep{shah2025comprehensive7}.\\
\indent With the widespread deployment of underwater pipelines, manual inspection not only consumes substantial human and material resources but also poses considerable safety risks~\citep{ioannou2024underwater13}.  Driven by the continuous integration of artificial intelligence, advanced sensing, and underwater robotics, the development of intelligent detection for underwater pipelines has emerged as a key research direction~\citep{rumson2021application14, bharti2022autonomous15}. Although numerous empirical and probabilistic intelligent detection models have been developed to assess structural failure~\citep{wang2024probabilistic8, yazdi2022operational9, ling2023data10}, the complex coupling of corrosion-related factors limits their ability to accurately capture the highly nonlinear degradation behavior of corroded pipelines~\citep{li2025machine11}. As a result, these methods often fail to accurately reflect the actual in-service condition of underwater pipelines. In contrast, direct acquisition of 3D geometric information from the pipeline surface offers a more intuitive and reliable foundation for corrosion detection and structural assessment~\citep{yao2025comprehensive12}. 
According to the sensing modality, existing underwater 3D imaging methods are generally classified into three categories: sonar-based acoustic imaging techniques~\citep{jaber20253d16, sun2025two17}, passive vision imaging techniques using cameras~\citep{levy2023seathru18}, and active vision imaging techniques based on structured light~\citep{lin2024method19}. Acoustic methods can reconstruct large-scale underwater environments through sound-wave propagation, which makes them particularly suitable for applications such as seabed mapping.  However, their inherently low spatial resolution limits their effectiveness for detailed local inspection~\citep{xu2025rotation20}.  Passive vision 3D imaging techniques primarily rely on camera to capture images, which are then reconstructed into 3D space through high-performance algorithms. These methods can perform well in bright and low interference environments. However, image degradation can be caused by light scattering and absorption in underwater environments~\citep{wang2024dual_add1,xu2025sfudnet_add2}, thereby reducing the reliability of 3D imaging algorithms.~\citep{fan2023structured21}. The active vision imaging techniques based on structured light combine laser emitter with camera to project structured light and capture its reflections for depth measurement. This method enables high-precision surface reconstruction even in dim and turbid environments while maintaining robustness~\citep{lin2025high22, simetti2020sea23, ye2025parallelization23.2}. It is particularly effective for detecting small-scale defects on the surface of pipelines. In recent years, structured light 3D imaging systems have achieved substantial progress in key technical areas such as system calibration, information fusion, image processing, and system integration. Recent studies by Ou et al.~\citep{ou2025underwater24, ou2024hybrid25} have demonstrated that the application of structured light technology to underwater detection achieves a favorable balance between spatial resolution and measurement accuracy, providing a reliable technical pathway for underwater pipeline detection tasks.\\
\indent  Most existing studies on structured light imaging have mainly concentrated on theoretical analysis, single-module optimization, or applications in fields other than pipeline detection~\citep{fan2023development_57}. As a result, research on system-level intelligent imaging for underwater pipeline detection remains limited. This limitation is further compounded by the fact that underwater pipelines are spatially extensive and exhibit diverse types of surface and structural defects. Existing structured light imaging systems for pipelines still lack an integrated framework that supports intelligent measurement, high-precision restoration, and adaptive robustness~\citep{palomer2019inspection26,fan2023structured21}. Consequently, it is difficult to maintain both accuracy and stability in complex underwater environments. To address these limitations, this paper presents a multi-mode underwater structured light imaging system for pipeline detection based on multi-source information fusion. The system adopts structured light active vision as the primary sensing module and integrates acoustic and inertial sensors for comprehensive detection. In particular, a deep learning-based pipeline edge detection network is incorporated to achieve intelligent pipeline detection in underwater environments. The proposed intelligent detection framework significantly enhances the overall performance and environmental adaptability of the UW-SLD system, thereby providing a solid foundation for long-term monitoring and maintenance of underwater pipelines. The main contributions of this work are summarized as follows:
\begin{itemize}
	\item A complete multi-mode structured light 3D imaging system for pipeline detection has been developed based on multi-source information fusion. It has made breakthroughs in multiple key technologies in the field of structured light imaging, such as rapid and high-precision system calibration, robust multi-source information fusion, intelligent image processing, and the multi-mode imaging strategy. 
	\item To compensate for nonlinear distortions caused by optical refraction in underwater imaging, a fast distortion correction method is proposed with high-precision and efficient. In addition, a factor graph based extrinsic calibration method between structured light scanner and doppler velocity log (DVL) is proposed to achieve unified coordinate alignment. 
	\item To enhance the overall detection accuracy in dynamic environments, a hierarchical multi-frequency fusion strategy is designed. On this basis, an AEKF method is proposed, which dynamically adjusts fusion weights according to sensor noise variations. This method substantially improves the accuracy and robustness of pose estimation.
	\item Building upon sensor calibration and information fusion, a multi-mode imaging strategy is proposed. Through the mechanism design and motion control, the system can flexibly support translation, rotation, and translation–rotation imaging modes. To enhance the data processing efficiency for these modes, a pipeline edge detection network is introduced to extract region of pipeline. It can suppress environmental interference and reduce redundant data. On this basis, an edge detection-based ICP algorithm is proposed to achieve robust and high-accuracy point cloud registration under variable speed motion. 
	\item To validate the effectiveness of the UW-SLD system, extensive experiments under varying operation modes, velocities, and depths are carried out. These experimental results demonstrate that the developed system can achieve superior accuracy, robustness, and adaptability. 
\end{itemize}
\par
\indent The rest of this paper is organized as follows. Section \ref{section2} summarizes the related research work; Section \ref{section3} elaborates on the overall hardware integration, software construction, and workflow of the system. Section \ref{section4} introduces the multi-sensor joint calibration method. Section \ref{section5} describes the multi-source information fusion method. Section \ref{section6} provides an introduction to the multi-mode structured light 3D imaging method. Section \ref{section7} illustrates the experimental setup, results, and interpretation of the results. \ref{section8} summarizes the work of this paper and looks forward.

\section{Related works}
\label{section2}
\subsection{Underwater pipeline detection methods}
\label{section2.1}
\indent Underwater pipeline detection can be divided into pipeline edge detection and anomalies detection such as surface defects and environmental risks. With the continuous advancement of underwater energy development and resource transportation, the demand for efficient and low-cost underwater pipeline detection methods is increasing rapidly~\citep{palomer2019inspection26, fuentes2025splash27}. In recent years, extensive research on underwater pipeline detection based on acoustic and optical imaging principles has driven the field into a stage of rapid development. Acoustic imaging-based detection methods can exploit the propagation characteristics of underwater sound waves to achieve long-range detection of large-scale targets~\citep{shi2024sonar28, wang2019deep28_2,li2025cross_add4, zhang2024bridge_56}. For instance, Zhang et al. realized pipeline edge detection and tracking in large-scale seabeds by multi-beam forward looking sonar~\citep{zhang2022submarine29}. Moreover, the forward looking sonar can be used to detect pipeline leaks and bubble emissions~\citep{zhang2021subsea30}. Kasetkasem et al. proposed a pipeline edge detection algorithm based on the self-organizing map using forward-looking sonar~\citep{kasetkasem2020pipeline31}. However, the high cost and limited resolution of acoustic imaging systems constrain their ability to reconstruct the geometric details and micro-defect structures on pipeline surfaces. Therefore, such methods are more suitable for macroscopic surveying of large-scale scenes, but are less effective for refined anomaly identification. In contrast, passive optical imaging methods primarily rely on cameras to detect and locate targets ~\citep{sang2023autonomous32, he2025ain_54, sun2024turbid_55}. Despite their cost efficiency, these methods are constrained by dim underwater illumination, light scattering, and attenuation during propagation, which hinder the acquisition of clear texture and depth information~\citep{zhao2025psnet_add3}. For instance, detection methods based on YOLO can only perform a coarse defect detection of underwater facilities (e.g., pipelines)~\citep{zhao2020research33, ye2024advanced_53}. Segmentation-based algorithms can restore the edge contours of the pipeline, but it is difficult to precisely depict the detailed defects on the pipeline surface~\citep{999808034}. Shen et al. have conducted research on polarized binocular 3D imaging technology based on image depth estimation~\citep{shen2025polarimetric34_2}. Although this method enables passive 3D imaging in turbid water, a comprehensive evaluation of its precision in capturing three-dimensional details is still lacking. In conclusion, acoustic and passive optical imaging methods have been partially applied in underwater pipeline detection. However, their applicability remains limited, and universal 3D imaging of diverse pipelines has not yet been achieved.\\
\indent Overall, the above-mentioned imaging methods have certain application value in specific scenarios, but they are still insufficient in dealing with complex and diverse pipelines. According to the service status of pipelines and their interactions with the surrounding environment, underwater pipeline anomalies can be classified into six types: leakage, depression, attachment, suspension, burial, and exposure~\citep{doi35}. For these diverse anomalies, the active visual imaging technology based on structured light can simultaneously obtain the fine geometric features of the pipeline surface and its spatial relationship with the surrounding environment. Therefore, it is more suitable for the detection and classification of different types of pipeline anomalies~\citep{hu2024novel36}. To better meet the imaging requirements of multiple scenarios and reduce system deployment costs, a multi-mode underwater structured light imaging system is proposed in this paper. This system integrates multi-source information fusion and deep learning methods on the basis of active vision to enhance robustness and intelligence. It provides a systematic solution for underwater pipeline intelligent detection and comprehensive anomaly identification.
\subsection{Key technologies of structured light 3D imaging system for underwater facilities}
\label{section2.2}
\indent The structured light is particularly suitable for 3D imaging of underwater facilities such as pipelines. It can effectively overcome the interference of the underwater dark environment and achieve precise imaging of the target surface. The key technologies of structured light imaging systems can be roughly divided into system calibration technology~\citep{zhang2024accurate37, zhao2022correction38, ou2023binocular39},  image processing technology~\citep{shi2025robust41, zhao2025robust42}, and specific system integration~\citep{gao2024underwater44, ikeda2024two45, fan2023structured46}. In these key technical fields, research efforts can be broadly classified into two categories: general technological development and system optimization tailored to specific application scenarios. For general technology research, Zhao et al. presented a general correction method for line structured light sensor~\citep{zhao2022correction38}. Ou et al. proposed an underwater refraction model and a multi-target laser calibration algorithm~\citep{ou2023binocular39}. There are also underwater image processing methods. Lin et al. proposed a method that converts underwater images into their aerial equivalents by integrating color enhancement with refraction distortion correction~\citep{lin2023conversion43}. Shi et al. employed a deep learning-based segmentation method to achieve laser stripe extraction~\citep{shi2025robust41}. In addition, a distance gradient regional energy method for underwater laser centerline extraction was proposed~\citep{zhao2025robust42}. In conclusion, the improvement of general imaging capabilities benefits diverse application scenarios and relies on the joint efforts of the research community. For specific imaging system integration, an advanced underwater concrete imaging system integrating structured light and stereo vision was developed by Lin et al.~\citep{lin2024method19, lin2025high22}. Additionally, there is a imaging system for nuclear fuel assembly detection developed by Feng et al.~\citep{feng2025underwater48}. For underwater pipeline detection, emphasis is placed on dynamic and multi-mode capabilities to meet diverse task requirements~\citep{palomer2019underwater49}. Therefore, this paper develops the UW-SLD system, which integrates these capabilities with multi-source information fusion to achieve high-accuracy and flexible underwater imaging for detection.

\section{Overview of UW-SLD system}
\label{section3}
\subsection{Hardware integration of the UW-SLD system}
To evaluate the system developed in this paper, the various hardware components are integrated. The hardware of the UW-SLD system can be divied into sensing and control components, as shown in \autoref{fig12} (a). The sensing part consists of dynamic laser, inertial vision and acoustic modules. The dynamic laser module has a high-frequency IMU, a laser, and a servo motor. It can achieve laser rotation via the high-precision servo motor, while vibration-induced errors are compensated using the high-frequency IMU. The inertial visual module integrates a camera and an IMU. This module can obtain the corresponding pose information while capturing the image. The acoustic module relies on a DVL as its core component to capture the pose of the system during movement. The combination of the dynamic laser module and the inertial vision module can realize the rotational structured light imaging. Furthermore, multi-mode imaging can be realized in dynamic scenes with the assistance of the acoustic module. The hardware of the control part consists of a master controller and two embedded controllers (ECs) for motion control. The master controller is used to execute the upper-level control strategies. One of the EC is used to control the dynamic laser module and obtain angular feedback. The another EC is used to control the experimental platform for realizing static, constant-speed, and variable-speed motion scenarios. In particular, the hardware connection in this paper is designed in a plug-and-play backpack form, allowing it to be directly mounted on various types of underwater robots. The \autoref{fig12} (a) shows the possible application carriers, including square-type, torpedo-type underwater robots, and even new types of bionic underwater robots.

\begin{figure*}[t!]
	\centering
	\subfigure[\label{fig12a}]{
		\includegraphics[width=7.5cm]{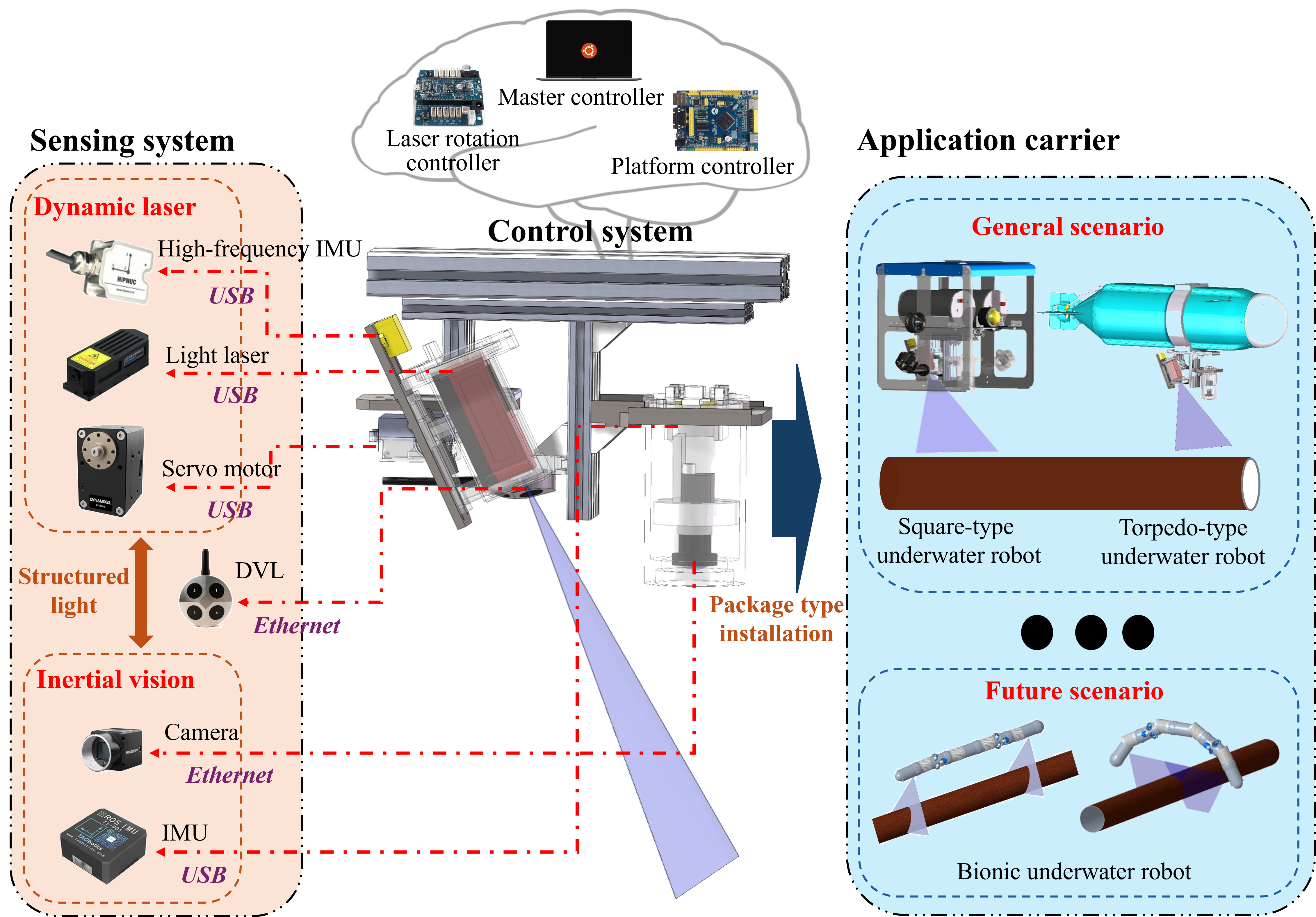}
	}
	\subfigure[\label{fig12b}]{
		\includegraphics[width=8cm]{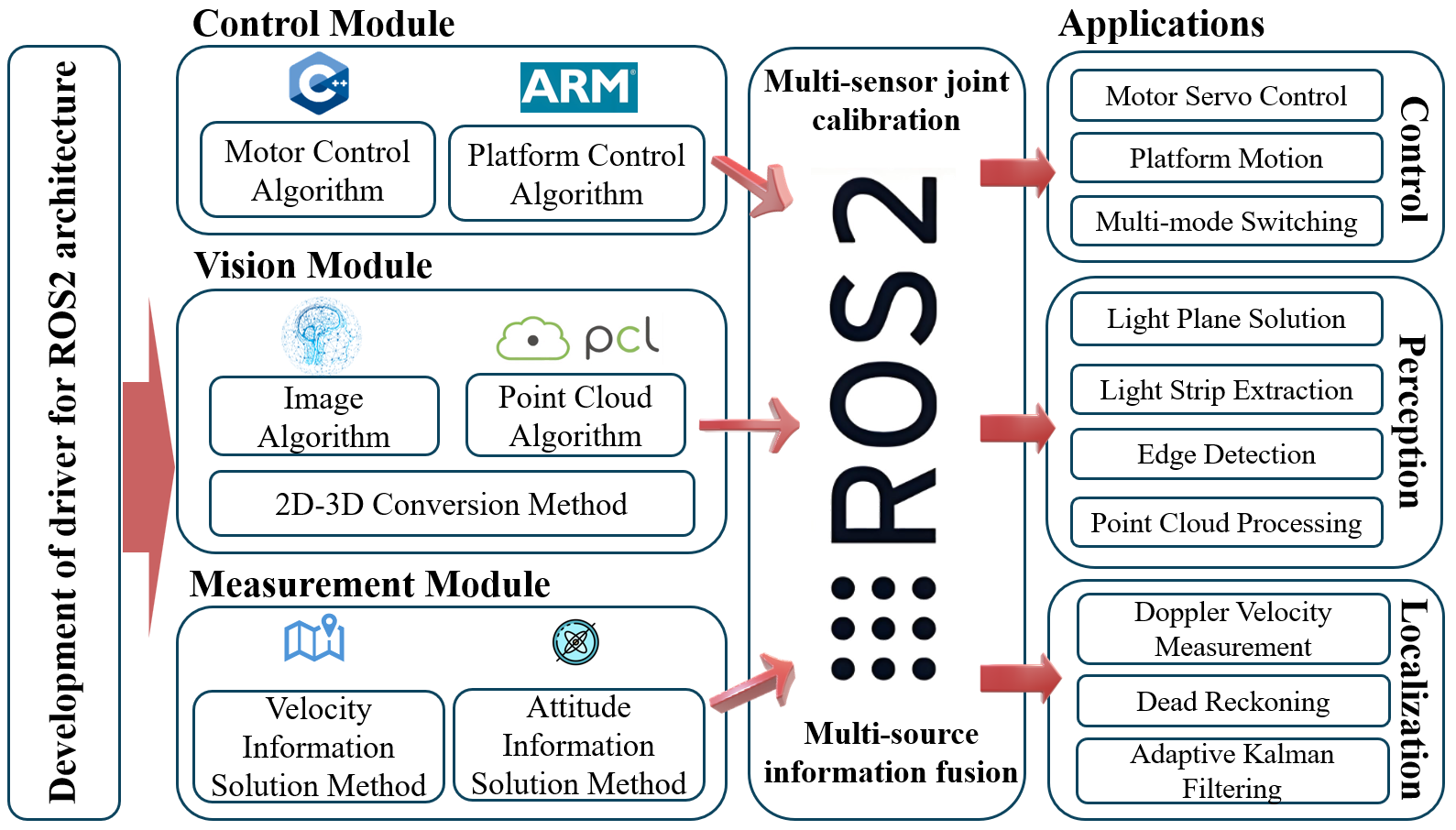}
	}
	\caption{Architecture of UW-SLD system. (a) Hardware integration of the UW-SLD system. (b) Software framework of the UW-SLD system.}
	\label{fig12}
\end{figure*}

\subsection{Software framework construction of the UW-SLD system}
The overall software framework is constructed on the ROS2 architecture, as illustrated in \autoref{fig12} (b). The framework adopts a modular and distributed design consisting of three primary software modules: control, vision, and measurement. The algorithms within all functional modules operate as independent nodes. The control module integrates both the driver and control algorithms for the servo motor and the motion platform. The vision module incorporates image, point cloud and 2D-3D conversion algorithms. The measurement module provides pose estimation through velocity and attitude information solution methods. Compared with ROS1, ROS2 exhibits significant advantages in system security, real-time communication, and cross-platform deployment. It is particularly suitable for complex robotic systems in dynamic or uncertain environments. At the system level, data from all modules are integrated and complemented through multi-sensor joint calibration and multi-source information fusion methods. This hierarchical integration strategy can enhance the spatial and temporal consistency of heterogeneous data streams, thereby improving overall sensing precision and operational robustness. Based on the above foundation, the software framework can support a wide range of applications across three functional domains. The control domain includes motor servo control, platform motion coordination, and multi-mode switching at the imaging process. The perception domain includes light plane solution, light strip extraction, edge detection and point cloud processing. The localization domain includes doppler velocity measurement, dead reckoning and adaptive Kalman filtering. Overall, this paper establishes a robust, scalable and real-time software framework. It can maintain efficient and stable operation in complex dynamic environments.

\subsection{UW-SLD system opration process} 
The opration process of the UW-SLD system comprises two sequential stages, as illustrated in \autoref{fig3}. The first stage is system startup, initialization and joint calibration. The second stage is underwater structured light-based 3D imaging process. The system adopts a multi-thread, modular architecture based on ROS2. Each function of the system is encapsulated as an independent subsystem. This design enhances security, maintainability, and real-time performance. During the first stage, the process begins with main power activation, followed by the initialization of the system environment and communication interfaces.  Then, all sensors are self-calibrated. Joint calibration is carried out among the camera, IMU and DVL. At this stage, an accurate initial state of the sensors is established to ensure that all hardware and software modules operate under well-calibrated conditions for subsequent processing. After system preparation and activation of all nodes, the second stage is launched. The workflow integrates four concurrently operating subsystems.  The laser rotation subsystem regulates the servo motor to execute the predefined imaging trajectory. The information fusion subsystem synchronizes data across sensors. It applies thresholding and traversal algorithms to find matching intervals, and organizes synchronized data into structured queues. This method mitigates time deviations caused by heterogeneous update rates or communication delays, ensuring temporal consistency across multi-source measurements. The AEKF subsystem processes IMU and DVL data to generate refined pose information. By estimating the noise parameters online, the filter gain is adaptively adjusted in real time to accommodate variations in motion state and sensor degradation. The point cloud generation subsystem first introduces a pipeline edge detection network to extract the target pipeline. Then, light stripe feature points are extracted from the images, and pose data are applied to perform rotational and translational transformations on these feature points. The resulting points are initially represented in the camera coordinate system and subsequently mapped to the world coordinate system. This process generates a 3D point cloud with temporal alignment and spatial consistency, suitable for downstream tasks such as environmental perception and reconstruction.

\section{Multi-sensor joint calibration}
\label{section4}
\subsection{Underwater refraction model}
\label{section4.1}
\indent The refraction will occur when light passes through different mediums in underwater scenes. In order to make the optical sensors operate normally underwater as they do in the air, underwater refraction model has been studied. To avoid complex calibration and potential errors of the laser emitter, the quartz lens is mounted with its plane surface perpendicular to the laser beam. This configuration minimizes the influence of refractive index variations, as shown in \autoref{fig4}. For the camera, the light reflected from the object undergoes two refractions as it passes through the water before reaching the imaging plane, as shown in \autoref{fig4}. By our previous work~\citep{li2023rapid_58}, the underwater refraction imaging model is equivalently transformed into a single-viewpoint model. By adjusting the distance between the camera lens and the inner surface of the quartz glass, the refraction error is minimized.  The object reflection point $P$ is projected onto virtual light center $O'_C$ through the virtual single-viewpoint model. The difference between the virtual light center $O'_C$ and the real  light center $O_C$ is defined as the light refraction error. The distance $d'_0$ between the virtual optical center and the outside surface of the glass and the optical refraction error $\Delta d$ can be calculated. The equation is as follows:
\begin{equation}
	\begin{aligned}
		d'_0 &= \tan \delta (d_0 \tan \alpha + d_1 \tan \beta), \\[3pt]
		\Delta d &= (d_0 + d_1) - \tan \delta (d_0 \tan \alpha + d_1 \tan \beta).
	\end{aligned}
\end{equation}

\begin{figure*}[ht]
		\centering
 		\includegraphics[width=16cm]{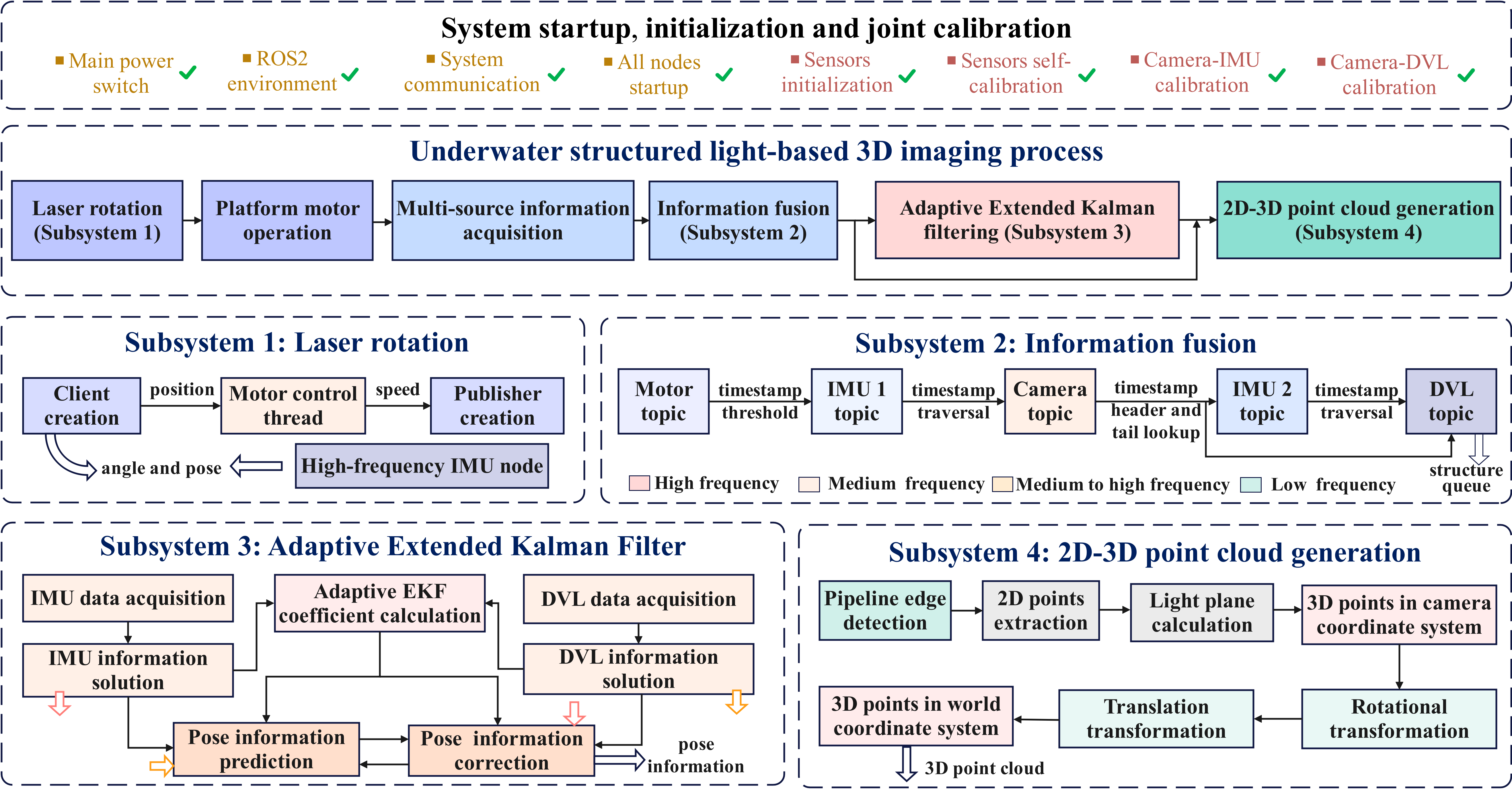}
	\caption{Operation process of UW-SLD system.\label{fig3}}
\end{figure*} 
\indent Further, the different image points are calculated to obtain the optimal $d_0$ when the $\Delta d$ range variation is the minimum using the least squares method.
\subsection{Underwater fast distortion correction}
\label{section4.2}

To address the optical distortion in underwater camera imaging, a fast and efficient distortion correction method is proposed. The detailed algorithm flow is presented in Algorithm \autoref{alg:undistort}, and its corresponding schematic diagram in \autoref{fig4}. The algorithm process is as follows:
\begin{figure}[H]
		\centering
		\includegraphics[width=7.5cm]{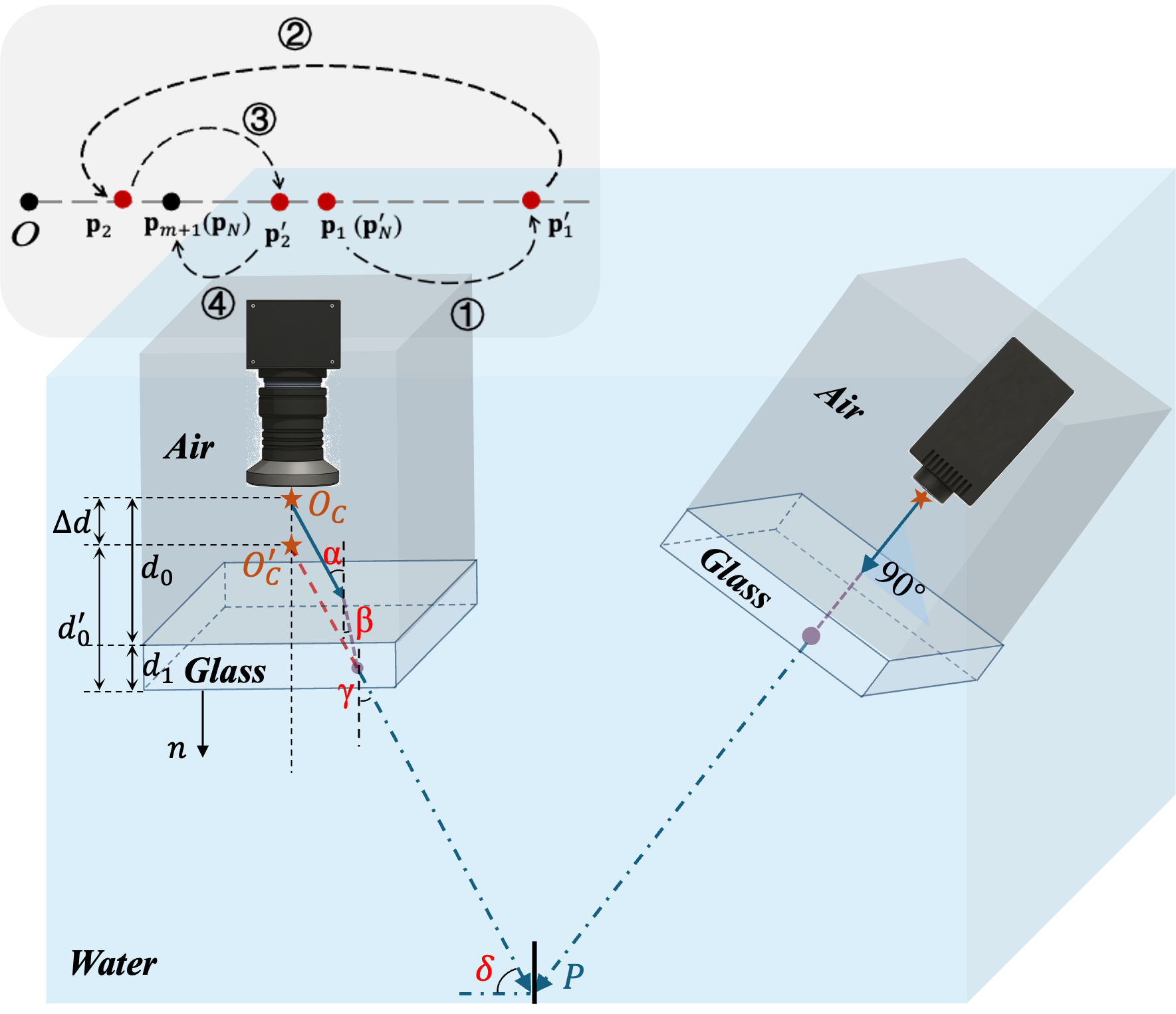}
	\caption{Underwater refraction principle diagram.\label{fig4}}
\end{figure}
\indent 1) Initialize all variables and set. Assume the observation point $\mathbf{p}'_N$ in the image corresponds to an undistorted point $\mathbf{p}_1$. Substitute it into the distortion function to obtain the corresponding distorted point $\mathbf{p}'_1$.
\begin{algorithm}[H]
	\caption{:Fast distortion correction}
	\label{alg:undistort}
	\footnotesize
	\begin{algorithmic}[1]
		\Require  distorted point $\mathbf p'_N \in \mathbb{R}^2$, distorted point set $\{\,{\mathbf p}'_m\,\}_{m=1}^{N}$, undistorted point set $\{\,{\mathbf p}_m\,\}_{m=1}^{N}$, image center $\mathbf O$, camera parameters $k_1,k_2,l_1,l_2,k_3$, iterations $m \in N$
		\Ensure  undistorted point $\mathbf p_N \in \mathbb{R}^2$
		\Function{Distortion}{$\mathbf p(x, y)$}
		\State $r^2 \gets x^2 + y^2$
		\State $p'_x \gets x\, (1+k_1 r^2+k_2 r^4+k_3 r^6) + 2l_1xy + l_2\,(r^2 + 2x^2)$ 
		\State $p'_y \gets y\,(1+k_1 r^2+k_2 r^4+k_3 r^6) + 2l_2xy + l_1\,(r^2 + 2y^2)$
		\State \Return $\mathbf p' \gets (p'_x, p'_y )^\top$
		\EndFunction
		\StepHeaderLen{0em}{Step 1: Initialization}
		\State Assuming undistorted point $\mathbf p_1 \gets \mathbf p'_N$
		\State Compute the distorted point $\mathbf p'_1 \gets \mathcal Distortion(\mathbf p_1)$
		\StepHeaderLen{0em}{Step 2: Stepping}
		\State $d_1 \gets \lVert  \mathbf p_1 -  \mathbf p'_1 \rVert$
		\State Radial unit vector $ \mathbf u \gets \dfrac{\mathbf p_N-\mathbf p'_N}{\lVert \mathbf p_N-\mathbf p'_N\rVert}$
		\State move from $\mathbf p_1$ by $d_1$ along $\mathbf u$: $\mathbf p^*_2 \gets \mathbf p_1 + d_1\,\mathbf u$
		
		\For{$m=2$ in $N$}
		\StepHeaderLen{.9em}{Step 3: Iterative update}
		\State Assuming undistorted point ${\mathbf p}_m \gets \mathbf p^*_m$
		\State Compute the distorted point $\mathbf{p}'_m \gets \mathcal
		Distortion({\mathbf p}_m)$
		
		\StepHeaderLen{.9em}{Step 4: Convergence check}
		\If{$\left|\overrightarrow{\mathbf p'_m \mathbf p'_N}\right|> \varepsilon$ }
	\State $d_m \gets \lVert  \mathbf p_m -  \mathbf p'_m \rVert$
	\State Radial unit vector $ \mathbf u \gets \dfrac{\mathbf p_N-\mathbf p'_N}{\lVert \mathbf p_N-\mathbf p'_N\rVert}$
	\State move from $\mathbf p_1$ by $d_m$ along $\mathbf u$: $\mathbf{p}^*_{m+1} \gets \mathbf p_1 + d_m\,\mathbf u$
	\Else
	\State \Return \textbf{$\mathbf p_N \gets \mathbf{p}^*_{m+1}$}
	\EndIf	
	
	\EndFor
	\State \Return \textbf{$\mathbf p_N \gets \mathbf{p}^*_{m+1}$}
\end{algorithmic}
\end{algorithm}
\indent 2) Considering that image distortion decreases toward the image center, the distortion degree of  $|\overrightarrow{\mathbf p_1 \mathbf p'_1}|$ is greater than $|\overrightarrow{\mathbf p_N \mathbf p'_N}|$. Starting from $\mathbf{p}'_N$, the $\mathbf{p}_2$ can be obtained move along $\overrightarrow{\mathbf p'_N \mathbf p_N}$ with a step length of $|\overrightarrow{\mathbf p_1 \mathbf p'_1}|$.\\
\indent 3) Based on the results from the first step, iterative updates are performed to correct the distortion.  Let the maximum number of iterations be $N$, and the current iteration index be $m$. The point $\mathbf{p}_2$ is again treated as an undistorted point. Then substitute this point into the distortion function to obtain the corresponding distorted point $\mathbf{p}'_2$. When point $\mathbf{p}_2$ lies between points
$\mathbf{p}_N$ and $\mathbf{p}_1$, it satisfies the constraint of 
$\left|\overrightarrow{\mathbf{p}_2 \mathbf{p}_2^{\prime}}\right|<\left|\overrightarrow{\mathbf{p}_N \mathbf{p}_N^{\prime}}\right|<\left|\overrightarrow{\mathbf{p}_1 \mathbf{p}_1^{\prime}}\right|$.  Starting from point $\mathbf{p}'_N$, point $\mathbf{p}_3$ can be obtained by moving along the $\overrightarrow{\mathbf p'_N \mathbf p_N}$  with a step length of $|\overrightarrow{\mathbf p_2\mathbf p'_2}|$. This process is then iteratively repeated.\\
\indent 4) The termination conditions such as the maximum number of iterations and threshold criteria are defined. Once any of these conditions is satisfied, the final undistorted point $\mathbf{p}'_N$ are obtained.
\subsection{Sensors extrinsic calibration across air-water medium}
The accurate IMU and camera extrinsic calibration is necessary for achieving high-precision imaging tasks. At present, numerous effective methods have been proposed for this calibration. In this paper, the Kalibr method~\citep{furgale2013unified_59} is employed to calibrate the transformation matrix between the camera and the IMU in air medium.\\

\begin{figure}[H]
		\centering
		\includegraphics[width=8cm]{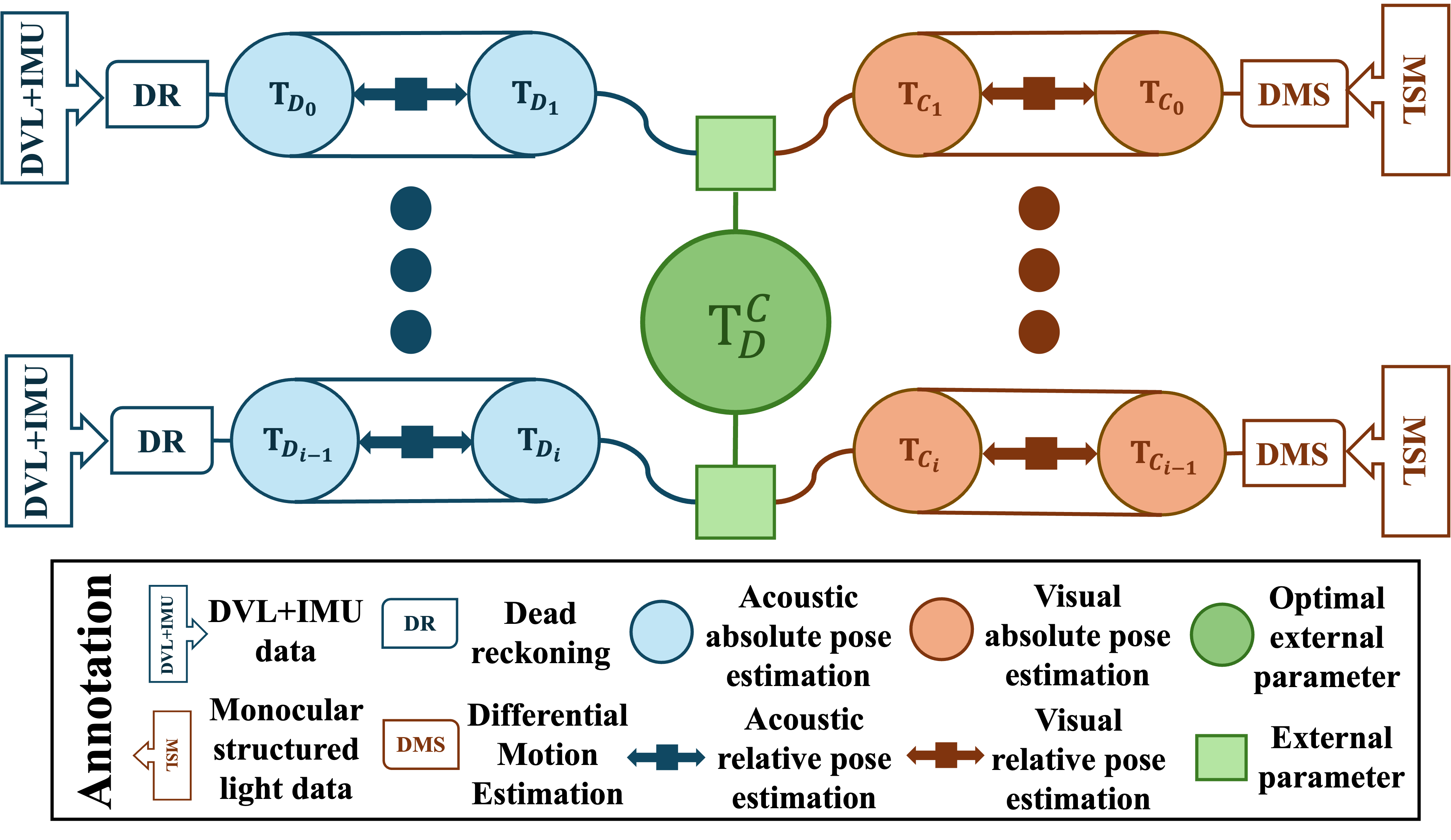}
	\caption{Schematic diagram of the external parameter calibration based on factor graph. \label{fig6}}
\end{figure} 
\indent In underwater media, vision and acoustic sensors cannot be extrinsically calibrated through nested installation as is done in vision–inertial calibration. To address this challenge, a factor-graph based extrinsic calibration method is proposed for the structured light camera and the DVL. The schematic diagram of the proposed method is shown in \autoref{fig6}. The specific process is as follows:
\begin{equation}
\mathbf{X}\cdot	\hat{\mathbf{T}}_{D_i}^{D_{i-1}}=\mathbf{T}_{C_i}^{C_{i-1}\cdot} \mathbf{X} 
\end{equation}
where, $\mathbf{X}$ denotes the ideal external transformation matrix between the structured light camera and the DVL. $\hat{\mathbf{T}}_{D_i}^{D_{i-1}}$ is the relative pose transformation  of the DVL from $i-1$ to $i$, which is estimated through the $\mathbf{X}$ and the $\mathbf{T}_{C_i}^{C_{i-1}}$. $ \mathbf{T}_{C_i}^{C_{i-1}}$ is the relative pose transformation of the structured light camera from $i-1$ to $i$. This pose transformation is calculated using the translation imaging method proposed in Section \ref{section6.2}. In underwater environments, these sensors are often disturbed by a large amount of noise. Therefore, the above ideal external parameter cannot be strictly satisfied. To solve this problem, an optimization model based on this ideal external parameter is constructed. The residual matrix $\mathbf{r}_i$ is constructed by comparing $\hat{\mathbf{T}}_{D_i}^{D_{i-1}}$ with $\mathbf{T}_{D_i}^{D_{i-1}}$, which is defined as follows:
\begin{equation}
	\mathbf{r}_i=\log \left((\mathbf{T}_{D_i}^{D_{i-1}})^{-1}\hat{\mathbf{T}}_{D_i}^{D_{i-1}}\right)^{\vee} \in \mathbb{R}^6
\end{equation}
where, $\log (\cdot)^{\vee}: S E(3) \rightarrow \mathfrak{s e}(3)$ is a mapping from the Lie group to the Lie algebra. The optimal external parameter estimation $\mathbf{X}^*=\mathbf{T}^{C}_{D}$ is obtained by minimizing the weighted sum of squared residuals at all times, as follows:
\begin{equation}
	\mathbf{X}^*=\arg \min _{\mathbf{X} \in S E(3)} \sum_{i=1}^N\left\|\mathbf{r}_i\right\|_{\Sigma_i}^2
\end{equation}
where, $\left\|\mathbf{r}_i\right\|_{\Sigma_i}^2=\mathbf{r}_i^T \Sigma_i^{-1}\mathbf{r}_i$ denotes Mahalanobis distance. $\Sigma_i$ is the covariance matrix. The above problem is iteratively solved through the Levenberg-Marquardt (LM) algorithm. In each iteration, an increment $\Delta x$ is solved in the tangent space of $\mathbf{X}$. The current estimated value is updated through exponential mapping, as follows:
\begin{equation}
	\mathbf{X}^* \leftarrow \mathbf{X} \cdot \exp \left(\Delta x^{\wedge}\right)
\end{equation}
where, $(\cdot)^{\wedge}$ is the inverse operation of $(\cdot)^{\vee}$. During the optimization process, the rotation matrix $\mathbf{R}_{\mathbf{X}^*}$ may slightly deviate from the orthogonality constraints of group $S O(3)$. Therefore, after the iteration terminates or necessary, the correction of the rotation matrix is performed. The correction is achieved through SVD orthogonalization, as described below:
\begin{equation}
	\begin{aligned}
		[\mathbf{U}, \mathbf{S}, \mathbf{V}] &= \operatorname{svd}\!\left(\mathbf{R}_{\mathbf{X}^*}\right), \\
		\mathbf{R}_{\mathbf{X}^*} &\leftarrow \mathbf{U}\mathbf{V}^\top.
	\end{aligned}
\end{equation}

\section{Multi-Source information fusion}
\label{section5}
\subsection{Hierarchical multi-frequency fusion strategy}
\indent The underwater structured light imaging system requires precise synchronization of multiple heterogeneous sensors operating at different frequencies. Therefore, a hierarchical multi-frequency fusion strategy is proposed. The schematic diagram is shown in \autoref{fig7}. The overall fusion strategy is divided into two layers: pose fusion and perceptual fusion. For the pose fusion layer, high-frequency information is collected from inertial sensor, while low-frequency information comes from acoustic sensor. The mid-frequency information is obtained by AEKF. This fusion process ensures smooth and reliable estimation of the nonlinear motion system, even in the presence of noise or partial information loss. For the perceptual fusion layer, high-frequency information is obtained from inertial sensor, mid-frequency information from servo feedback, and low-frequency information from camera. The integration of these sensors can compensate for the measurement errors and drift of the sensors caused by oscillation during the imaging process. The fused information from the both layers is subsequently synchronized with the structured light imaging system. This ensures the temporal alignment between motion estimation and 3D point cloud generation.
\begin{figure}[H]
		\centering
		\includegraphics[width=8cm]{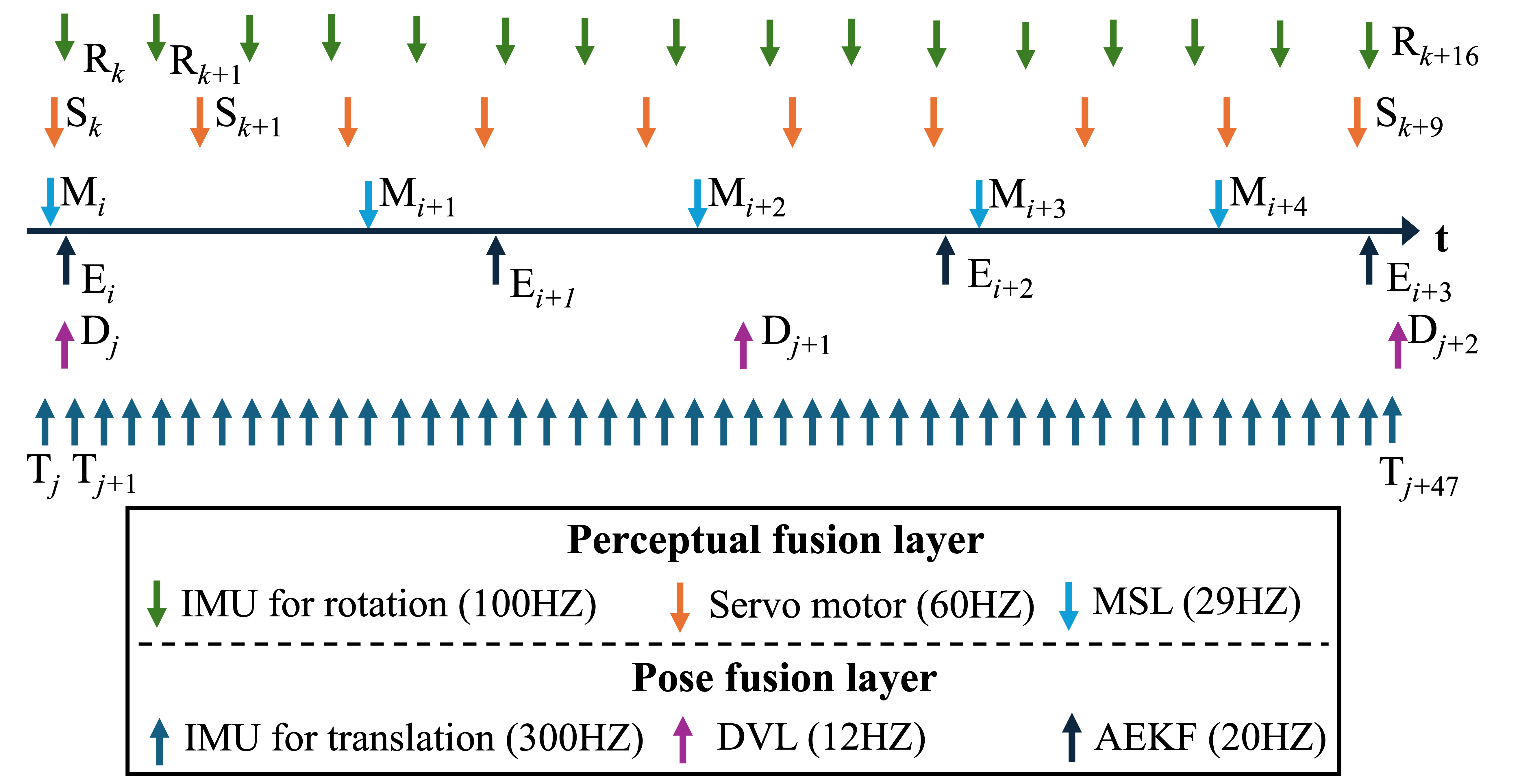}
	\caption{Hierarchical multi-frequency fusion strategy diagram.\label{fig7}}
\end{figure} 
\subsection{Inertial - Acoustic pose fusion based on AEKF}
The hierarchical fusion strategy is built on a frequency-domain complementary framework. It leverages the intrinsic characteristics of different sensors. For the pose fusion layer, robust pose estimation is achieved through high-frequency inertial prediction, low-frequency acoustic correction, and mid-frequency  filtering fusion. \\
\indent 1) High-frequency inertial prediction: The point cloud pose in the world coordinate system is obtained by performing numerical integration on the high-frequency IMU data. The discrete-time kinematic propagation of position $\mathbf{p}_k$, velocity $\mathbf{v}_k$, and attitude $\mathbf{q}_k$ can be expressed as follows:
\begin{equation}
	\begin{aligned}
		\mathbf{p}_k &= \mathbf{p}_{k-1}+\mathbf{v}_{k-1}\Delta t_k 
		+ \tfrac{1}{2}\mathbf{R}(\mathbf{q}_{k-1})\mathbf{a}_{k-1}\Delta t_k^2 
		+ \tfrac{1}{6}\mathbf{j}_k\Delta t_k^3,\\
		\mathbf{v}_k &= \mathbf{v}_{k-1}+\mathbf{R}(\mathbf{q}_{k-1})\mathbf{a}_{k-1}\Delta t_k,\\
		\mathbf{q}_k &= \mathbf{q}_{k-1}
		+ \tfrac{1}{2}(\mathbf{q}_{k-1}\otimes\boldsymbol{\omega}_{k-1})\Delta t_k
		+ \tfrac{1}{4}(\mathbf{q}_{k-1}\otimes\boldsymbol{\alpha}_k)\Delta t_k^2.
	\end{aligned}
	\label{equ8}
\end{equation}

where, $k$ denotes the sampling index, and $\Delta t_k$ represents the sampling interval between time steps $k-1$ and $k$.  $\mathbf{j}_k$ is the linear jerk vector, defined as the time derivative of the linear acceleration. $\boldsymbol{\alpha}_k$ denotes the angular acceleration vector. $\otimes$ denotes the quaternion multiplication operator.\\
\indent 2) Low-frequency acoustic correction: The low-frequency velocity information from DVL is used as an absolute state constraint. It can correct the long-term accumulated errors of the IMU in the underwater environment. Therefore, the velocity and position calculation in \autoref{equ8} has been updated to: 
\begin{equation}
	\setlength{\jot}{1.5pt} 
	\begin{aligned}
		\mathbf{v}_k &=(1-\lambda)\big(\mathbf{v}_{k-1}
		+\mathbf{R}(\mathbf{q}_{k-1})\,\mathbf{a}_{k-1}\Delta t_k\big)+ \lambda\, \mathbf{R}_I^{D}\,\mathbf{v}_k^{D},\\[2pt]
		\mathbf{p}_k &=(1-\lambda)\big(\mathbf{p}_{k-1}
		+\mathbf{v}_{k-1}\Delta t_k
		+\tfrac12 \mathbf{R}(\mathbf{q}_{k-1})\,\mathbf{a}_{k-1}\Delta t_k^{2}
		\\ &\quad 	+\tfrac16 \mathbf{j}_k \Delta t_k^{3}\big)+ \lambda\, \mathbf{T}_I^{D}\,\mathbf{p}_k^{D}.
	\end{aligned}
\end{equation}

where, $\lambda$ represents the health status of DVL. If $\lambda$ is verified to be healthy, then $\lambda=1$; otherwise, $\lambda=0$. $\mathbf{T}_I^D$ denotes the transformation matrix from the DVL coordinate system to the IMU coordinate system.
\begin{algorithm}[H]
	\caption{: Adaptive Extended Kalman Filtering}
	\label{alg:AEKF}
	\footnotesize
	\begin{algorithmic}[1]
		\State \textbf{Input:}  error matrices $\mathbf{Q}$, $\mathbf{R}$, threshold $K_m$, initial state $\mathbf{x}_{k-1}$ and covariance matrix $\mathbf{P}_{k-1}$
		\State \textbf{Output:} $\hat{\mathbf{x}}_{k}$
		\State \textbf{Given} nonlinear system state space model:
		\Statex \hspace{\algorithmicindent} $\mathbf{x}_k = f(\mathbf{x}_{k-1}, \mathbf{u}_{k-1}, \mathbf{w}_{k-1}), \;\; \mathbf{w}_{k-1}\sim\mathcal{N}(\mathbf{0},\mathbf{Q})$
		\Statex \hspace{\algorithmicindent} $\mathbf{z}_k = h(\mathbf{x}_k,\mathbf{v}_k), \;\; \mathbf{v}_k\sim\mathcal{N}(\mathbf{0},\mathbf{R})$
		\State \textbf{Linearization}:
		\Statex \hspace{\algorithmicindent} $\mathbf{x}_k = f(\hat{\mathbf{x}}_{k-1}, \mathbf{u}_{k-1}, \mathbf{w}_{k-1}) +\mathbf{A}(\mathbf{x}_k-\hat{\mathbf{x}}_{k-1})+ \mathbf{G} \mathbf{w}_{k-1}$
		\Statex \hspace{\algorithmicindent} $\mathbf{z}_k = h(\hat{\mathbf{x}}_k^{-},\mathbf{v}_k)+\mathbf{H}(\mathbf{x}_k-\hat{\mathbf{x}}_k^{-})+\mathbf{V}\mathbf{v}_k$
		\For{each time step $k$}
		\State Compute Jacobians: $\mathbf{A}$, $\mathbf{G}$, $\mathbf{H}$, $\mathbf{V}$
		\State Prior state prediction: $\hat{\mathbf{x}}_k^{-} = f(\hat{\mathbf{x}}_{k-1}, \mathbf{u}_{k-1}, \mathbf{w}_{k-1})$
		\State Prior covariance prediction: $\mathbf{P}_k^{-} = \mathbf{A} \mathbf{P}_{k-1} \mathbf{A}^\top + \mathbf{G} \mathbf{Q} \mathbf{G}^\top$
		\State Innovation: $\tilde{\mathbf{y}}_k = \mathbf{z}_k - h(\hat{\mathbf{x}}_k^{-})$
		\State Kalman gain: $\mathbf{K}_k = \mathbf{P}_k^{-} \mathbf{H}_k^\top (\mathbf{H}_k \mathbf{P}_k^{-} \mathbf{H}_k^\top + \mathbf{V}\mathbf{R}\mathbf{V}^\top)^{-1}$
		\State State update: $\hat{\mathbf{x}}_k = \hat{\mathbf{x}}_k^{-} + \mathbf{K}_k \tilde{\mathbf{y}}_k$
		\State Covariance update: $\mathbf{P}_k = (\mathbf{I} - \mathbf{K}_k \mathbf{H}_k) \mathbf{P}_k^{-}$
		\State Actual innovation covariance: $\mathbf{P}_{a,k} = \tilde{\mathbf{y}}_k \tilde{\mathbf{y}}_k^\top$
		\State Predicted innovation covariance: $\mathbf{P}_{e,k} = \mathbf{H}_k \mathbf{P}_k^{-} \mathbf{H}_k^\top + \mathbf{V}\mathbf{R}\mathbf{V}^\top$
		\State Diagonalize: $\mathbf{D}_{a,k} = \operatorname{diag}(\mathbf{P}_{a,k}), \quad \mathbf{D}_{e,k} = \operatorname{diag}(\mathbf{P}_{e,k})$
		\State Compute adaptive parameter: $K_n = \operatorname{Trace}(\mathbf{D}_{a,k} / \mathbf{D}_{e,k})$
		\If{$K_n > K_m$}
		\State $\eta_k = 1 / (K_n + 1\text{e}^{-8}) - 1$
		\ElsIf{$0.1 < K_n < K_m$}
		\State $\eta_k = K_n$
		\Else
		\State $\eta_k = 0$
		\EndIf
		\State Adjust covariance: $\mathbf{P}_k \gets (\mathbf{I} - \mathbf{K}_k \mathbf{H}_k)\mathbf{P}_k^{-} - \eta_k \mathbf{K}_k \mathbf{H}_k \mathbf{P}_k^{-}$
		\EndFor
	\end{algorithmic}
\end{algorithm}

\indent 3) Mid-frequency filtering fusion: In underwater environments, the estimation of velocity, position and attitude often relies on the complementary information among multiple sensors. Therefore, the high-frequency inertial data and  low-frequency acoustic measurements are deeply fused by AEFK in this paper. The algorithm details are shown in Algorithm \autoref{alg:AEKF}. The acoustic measurements are treated as observation inputs and inertial data as prediction inputs. Initially, the timestamps of both sensor data streams are synchronized. Based on this alignment, complementary fusion is performed according to the error status of each sensor. Compared with the EKF, the AEKF introduces an adaptive factor $\eta_k$. It can achieve even more precise error compensation when complex environmental interference leads to sensor latency.
\section{Multi-Mode structured light 3D imaging method}
\label{section6}

\subsection{Multi-Mode structured light 3D imaging strategy}
\indent Due to the complex underwater terrain, the laying of underwater pipelines may present diverse geometric structures. This makes it difficult for a single imaging mode to achieve comprehensive coverage detection. To address this challenge, a multi-mode structured light imaging strategy is proposed. It can flexibly support translation, rotation, and translation–rotation imaging modes according to specific detection environments. For pipelines with regular arrangements, continuous 3D imaging is achieved through translation mode. When the pipeline is tilted or partially occluded, rotational imaging is employed to compensate for the loss of angular information and to minimize blind regions. For irregularly installed pipelines, the combined mode is adopted to ensure full coverage of the structured light across the target surface. \\
\indent The accurate estimation of the light plane is a prerequisite for implementing the aforementioned multi-mode imaging. Therefore, the estimation method has been intensively investigated in our previous work~\citep{hu2024novel36}. The specific algorithm for calculating the light plane is shown in Algorithm \autoref{alg:estimation}. The schematic diagrams of each variable are shown in \autoref{fig8}. For ease of distinction, $O_c-X_cY_cZ_c$ represents the camera coordinate system and $C$ represents a point in the algorithm process. Beyond the algorithm process, $O_C-X_CY_CZ_C$ represents the camera coordinate system and $O_W-X_WY_WZ_W$ represents the world coordinate system.
\begin{figure}[H]
		\centering
		\includegraphics[width=8cm]{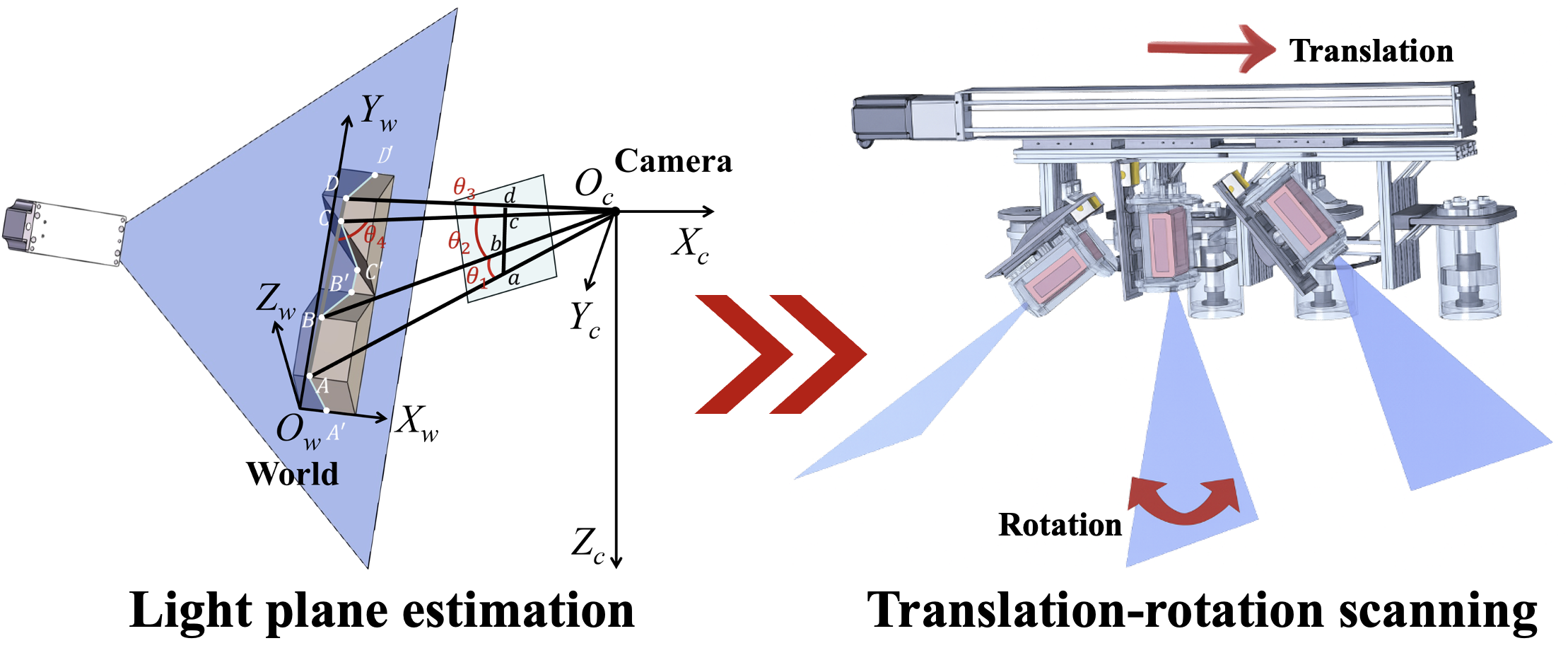}
	\caption{Schematic diagram of multi-mode structured light imaging.\label{fig8}}
\end{figure} 
\subsection{Underwater translational 3D imaging method}
\label{section6.2}
\indent The 3D points in the camera coordinate system are defined as the intersection of the imaging ray and the structured light plane. To enable translational imaging, the pose of the light plane is first calculated based on its parameters. Then, the rotation matrix of the light plane between the camera and world coordinate systems is obtained by singular value decomposition (SVD) on the basic transformation matrix. Based on this result, the translation matrix during the imaging process can subsequently be determined:
\begin{algorithm}[H]
	\caption{Structured light plane estimation}\label{alg:estimation}
	\footnotesize
	\begin{algorithmic}[1]
		\State \textbf{Input:}  pixel sampling points $(u_a, v_a)$, $(u_{a'}, y_{a'})$, $(u_b, y_b)$, $(u_{b'}, y_{b'})$, $(u_c, y_c)$, $(u_{c'}, y_{c'})$, $(u_d, y_d)$, $(u_{d'}, y_{d'})$; Y-axis coordinates $Y_A$, $Y_B$, $Y_D$, $Y_{A'}$, $Y_{B'}$, $Y_{D'} $ in $O_w-X_wY_wZ_w$ 
		\State \textbf{Output:}  light plane $\mathbf{\pi}_c=(a_c, b_c, c_c, d_c)^T$ in $O_c-X_cY_cZ_c$ , light plane $\mathbf{\pi}_w=(a_w, b_w, c_w, d_w)^T$ in $O_w-X_wY_wZ_w$. 
		\StepHeaderLen{0em}{Step 1: Locate $C$ and $C'$ via cross-ratio}\\
		$r \leftarrow \dfrac{|\overrightarrow{CA}|}{|\overrightarrow{DA}|} \cdot \dfrac{|\overrightarrow{CB}|}{|\overrightarrow{DB}|}
		= \dfrac{|\overrightarrow{ca}|}{|\overrightarrow{da}|} \cdot \dfrac{|\overrightarrow{cb}|}{|\overrightarrow{db}|}$\\
		$Y_C \leftarrow \dfrac{(Y_B-Y_A)(Y_D-Y_B)}{r(Y_D-Y_A)-(Y_D-Y_B)} + Y_B$\\
		$X_C \leftarrow$ intersection computed from edge slope of $C$\\
		Repeat the above for $C'$ to obtain $(X_{C'},Y_{C'})$
		
		\StepHeaderLen{0em}{Step 2: Compute inter-ray angles}\\
		$\theta_1 \leftarrow \arccos(\hat{\mathbf{o}}_{cA}\cdot\hat{\mathbf{o}}_{cB})$,\;
		$\theta_2 \leftarrow \arccos(\hat{\mathbf{o}}_{cB}\cdot\hat{\mathbf{o}}_{cC})$,\;
		$\theta_3 \leftarrow \arccos(\hat{\mathbf{o}}_{cC}\cdot\hat{\mathbf{o}}_{cD})$\\
		$k \leftarrow \dfrac{|\overrightarrow{BD}|}{|\overrightarrow{BC}|}\cdot\dfrac{\sin\theta_2}{\sin(\theta_2+\theta_3)}$\\
		$\theta_4 \leftarrow \operatorname{atan2}\!\big(-k\sin\theta_3,\; 1 - k\cos\theta_3\big)$
		
		\StepHeaderLen{0em}{Step 3: Solve point coordinates via sine theorem}\\
		$\gamma \leftarrow \dfrac{\sin(\pi-\theta_1+\theta_2+\theta_4)}{\sin(\theta_1+\theta_2)}$\\
		$\overrightarrow{O_cA} \leftarrow \dfrac{\sin \theta_4 \cdot |\overrightarrow{AC}| \cdot \hat{\mathbf{o}}_{cA}}{\sin \left(\theta_1+\theta_2\right)}  $,
		$\overrightarrow{O_cB} \leftarrow  \dfrac{\gamma \sin\theta_4 \cdot|\overrightarrow{AC}| \cdot \hat{\mathbf{o}}_{cB}}{\sin(\pi-\theta_2-\theta_4)}$\\
		$\overrightarrow{O_cC} \leftarrow \gamma\cdot | \overrightarrow{AC}| \cdot \hat{\mathbf{o}}_{cC}$,
		$\overrightarrow{O_cD} \leftarrow  \dfrac{\gamma\sin(\pi-\theta_4)}{\sin(\theta_4-\theta_3)}\cdot| \overrightarrow{AC}|\cdot \hat{\mathbf{o}}_{cD}$ \\
		Repeat for $|\overrightarrow{A'C'}|$ to obtain $\overrightarrow{O_cA'}$, $\overrightarrow{O_cB'}$, $\overrightarrow{O_cC'}$, $\overrightarrow{O_cD'}$
		\StepHeaderLen{0em}{ Step 4: Set geometric constraints}\\
		$\left|\overrightarrow{C C^{\prime}}\right|=\left|\overrightarrow{O_c C^{\prime}}-\overrightarrow{O_c C}\right|$, $\frac{|\overrightarrow{B D}|}{\left|\overrightarrow{B^{\prime} D^{\prime}}\right|}=\frac{\left|\overrightarrow{O_C D}-\overrightarrow{O_C B}\right|}{\left|\overrightarrow{O_C D^{\prime}}-\overrightarrow{O_C B}\right|}=\frac{Y_D-Y_B}{Y_{D^{\prime}}-Y_{B^{\prime}}}$
		
		\StepHeaderLen{0em}{ Step 5: Recover 3D points and fit plane $\pi_c$}\\
		Combining Step 3 and 4 for $|\overrightarrow{AC}|$ and $|\overrightarrow{A'C'}|$ \\
		Select three non-collinear points from $\{\mathbf{P}_A^c, ..., \mathbf{P}_D^c, ..., \mathbf{P}_{D'}^c\}$ to fit $\mathbf{\pi}_c: a_cx+b_cy+c_cz+d_c=0$
		
		\StepHeaderLen{0em}{ Step 6: Fit plane $\mathbf{\pi}_w$}\\
		Given $\mathbf{P}_C^w$, $\mathbf{P}_C'^w$ and the distances between each point, $\{\mathbf{P}_A^w, ..., \mathbf{P}_D^w, ..., \mathbf{P}_{D'}^w\}$ are calculated using the Pythagorean theorem\\
		Select three non-collinear points to fit
		$\mathbf{\pi}_w$ \\
		\Return $(\mathbf{\pi}_c, \mathbf{\pi}_w)$
	\end{algorithmic}
\end{algorithm}
\begin{equation}
	\mathbf{t}=\left[\begin{array}{c}
		X_W \\
		Y_W \\
		Z_W
	\end{array}\right]-\mathbf{R}_W^C\left[\begin{array}{c}
		X_C\\
		Y_C \\
		Z_C
	\end{array}\right]
\end{equation}
\indent By obtaining the translation vectors of two consecutive frames and their corresponding timestamps $\tau$, the inter-frame displacement matrix can be derived:
\begin{equation}
	\Delta \mathbf{t}=\frac{\mathbf{t}_2-\mathbf{t}_1}{\tau_2-\tau_1}
\end{equation}
\indent Based on the above results, the transformation matrix of the 3D points from the camera coordinate system to the world coordinate system can be obtained in \autoref{e12}. Finally, translational imaging is implemented by transforming the current 3D point using the corresponding transformation matrix.
\begin{equation}
	\label{e12}
	\mathbf{T}(\tau)=\left[\begin{array}{cc}
		\mathbf{R}_W^C & \Delta \mathbf{t} \cdot\Delta\tau+\mathbf{t} \\
		0 & 1
	\end{array}\right]
\end{equation}
\subsection{Underwater translation-rotation 3D imaging method}
\indent The translation-rotation imaging is achieved by coupling the translation imaging and the rotation imaging. Firstly, the pre-rotation plane $\mathbf{\pi}_1$ and the post-rotation plane $\mathbf{\pi}_2$ in the camera coordinate system need to be calibrated. The unit vector $\mathbf{d}_n$ of rotation axis is calculated using the normal vectors of these two planes. Then, the current light plane is calculated using the Rodrigue rotation theorem in real time. The Rodrigues rotation is an excellent scheme to handle rotation by any angle about any axis, and the equation is:
\begin{equation}
	\label{eq8}
	\mathbf{R}_n=\mathbf{I}+\sin \theta_r \cdot \mathbf{K}+(1-\cos \theta_r) \cdot \mathbf{K}^2
\end{equation}
where, $\theta_r$ is the rotation angle from the feedback of the servo motor. $\mathbf{K}$ denotes the antisymmetric matrix of the $\mathbf{d}_n$.\\
\indent Based on the above calculation, the initial plane $\mathbf{\pi}_1$ can be transformed into a new plane $\pi_r$ by a rotation of $\theta_r$ around the axis $\mathbf{d}_n$. Accordingly, the normal vector of the new plane can be represented as:
\begin{equation}
	\mathbf{n}_{\mathrm{\mathbf{\pi}_r}}=\mathbf{R}_n \cdot \mathbf{n}_{\mathbf{\pi}_1}=\left[a_r, b_r, c_r\right]^T
\end{equation}
\indent To calculate the new plane equation, a point $\mathbf{p}_r=(x_r, y_r, z_r)$ is selected on the rotation axis with $z_r = 0$. By combining the equations of planes $\pi_1$ and $\pi_2$, the coordinates of point $\mathbf{p}_r$ can be calculated. Then, the parameter $[a_r, b_r, c_r, d_r]$ can be obtained.\\
\indent The translation-rotation imaging can be applied to both uniform-speed and variable-speed scenarios. The displacement changes in different scenarios are as follows:
\begin{equation}
	\mathbf{t}(\tau)=\left\{\begin{array}{lll}
		\Delta \mathbf{t} \cdot\Delta\tau, & \text { if } \ \mu=0 \\
		\mathbf{R}_C^A \cdot \mathbf{V}_A(\tau) \cdot \Delta \tau, & \text { if } \mu=1 
	\end{array}\right.
\end{equation}
where, $\mu$ is a symbol. If $\mu=0$, the system is in a uniform-speed motion. When $\mu=1$, the system is in a variable-speed motion. $\mathbf{R}^A_C=\mathbf{R}^I_C$ denotes the transformation matrix from the AEKF coordinate system to the camera coordinate system.  $\mathbf{V}_A(\tau)$ represents the speed information obtained by fusing IMU and DVL data through AEKF. During the translation-rotation imaging, the transformation matrix can be updated to:
\begin{equation}
	\mathbf{T}(\tau)=\left[\begin{array}{cc}
		\mathbf{R}_W^C &  \mathbf{t}(\tau) \\
		0 & 1
	\end{array}\right]
\end{equation}
\subsection{Intelligent pipeline edge detection and efficient registration}
To achieve high-performance 3D imaging of underwater pipelines under variable-speed condition, this paper introduces an intelligent edge detection network from our previous work~\citep{hu2025context_52}, called CEPDNet.  The schematic diagram of the detection network is shown in \autoref{fig9}. A context-enhanced strategy is adopted in the network to achieve high-precision detection of keypoints along pipeline edges.  This network can effectively remove non-target regions from complex backgrounds. Moreover, it can improve imaging efficiency and reduce redundant data storage during underwater operations.  Based on the synergistic optimization of edge detection and point cloud registration, the ED-ICP algorithm is further proposed. The algorithm flow is shown in \autoref{fig10}. During the registration stage, ED-ICP achieves spatial structure optimization through edge detection. Subsequently, the pre-alignment is achieved through the pose guidance based on AEKF. Then, the kd-tree based nearest neighbor search and a weighted transformation matrix optimization are incorporated into the iterative process. It enables fast and stable convergence even under non-uniform point cloud densities. In addition, a distance transform method is employed to adaptively determine the maximum correspondence threshold based on the overlap rate. This method can effectively remove outliers and improve computational efficiency and accuracy. Through the above improvements, the system can achieve high-accuracy alignment and global merging of point clouds in imaging process. 
\begin{figure}[H]
		\centering
		\includegraphics[width=7cm]{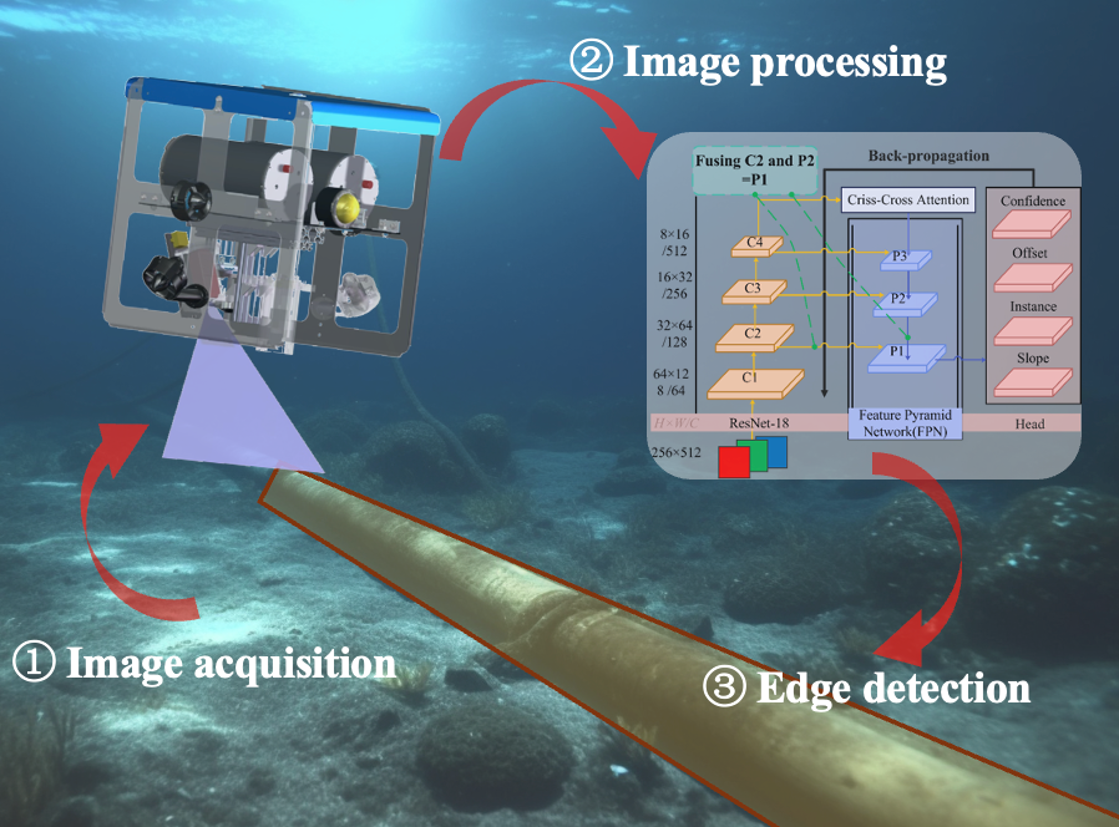}
	\caption{Schematic diagram of CEPDNet network for detecting pipeline edges.\label{fig9}}
\end{figure}
\section{Experiments}
\label{section7}
\subsection{Experimental setup}
\indent A series of experiments are conducted to evaluate the feasibility and performance of the developed system. The experiments are performed in a water pool with dimensions of 1.31 × 0.80 × 0.63 m. To achieve realistic underwater conditions, the pool are filled with a mixture of fine (0.5–1 mm), medium (2–3 mm), and coarse (4–6 mm) sand particles at ratio of 1:1:1. The experimental environment is shown in \autoref{fig11}. The experiments are divided into shallow water test and deep water test. The shallow water condition is defined with the camera positioned 0.36–0.39 m above the pool bottom. The deep water condition corresponded to a camera height of 0.50–0.55 m. During the constant speed test, the imaging platform moved steadily at approximately 3 mm/s. In the variable speed test, the initial velocity is 3 mm/s, followed by irregular random fluctuations to emulate real underwater motion. Before introducing the intelligent pipeline edge detection network, the system operated at a frequency of 23 Hz. After integration, the operating frequency increased to 29 Hz. This demonstrates a notable improvement in imaging efficiency.
\begin{figure}[H]
		\centering
		\includegraphics[width=8cm]{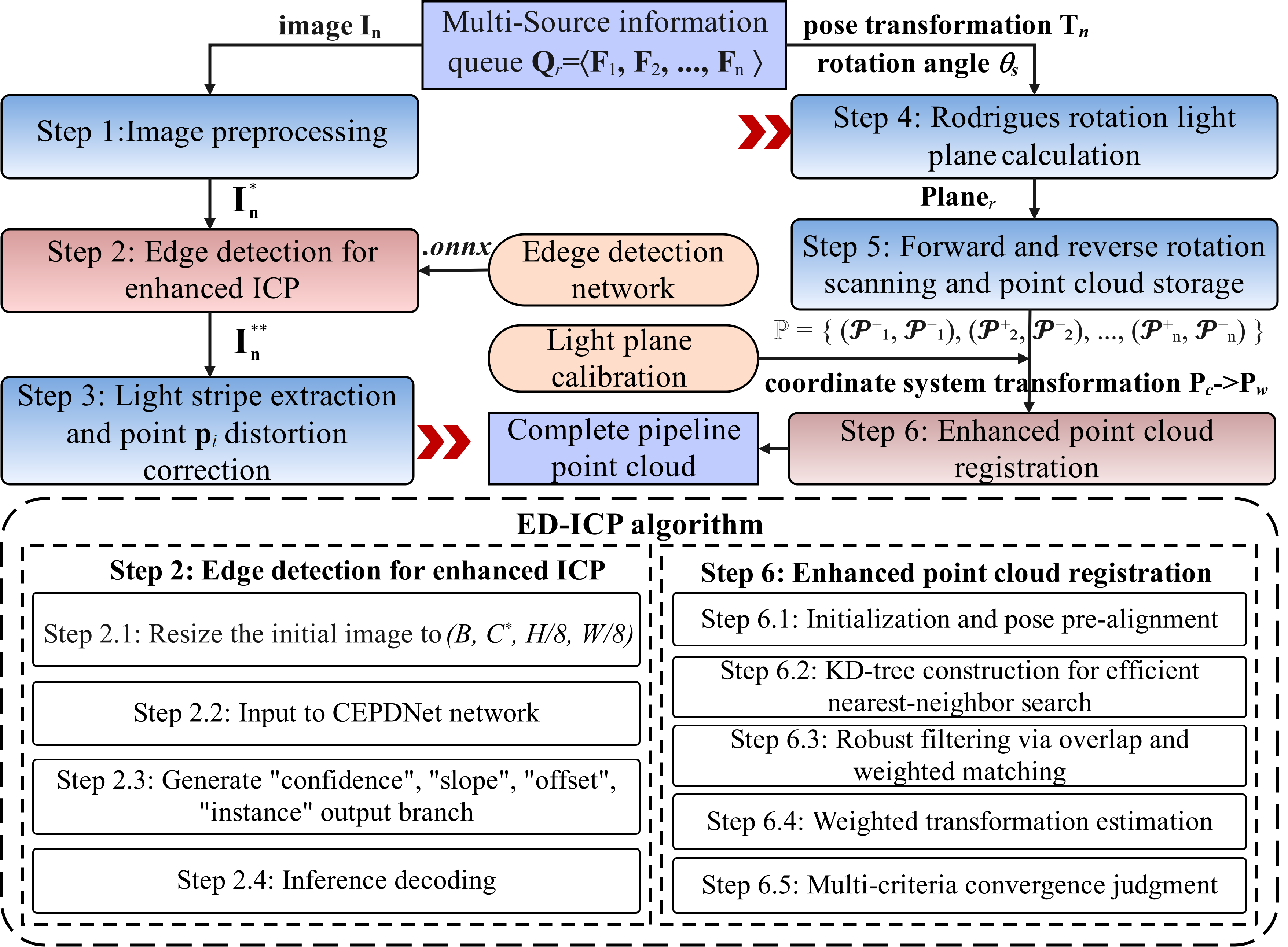}
	\caption{Flow of the ED-ICP algorithmn.\label{fig10}}
\end{figure} 
\indent The master controller is built on Ubuntu 20.04, and its software is developed based on the ROS2 framework. The onboard processing unit of the controller is equipped with an Intel (R) Core (TM) i7-12700H (2.7GHz) and an NVIDIA GeForce RTX 3050 Ti (4GB). The edge detection model is trained on a high-performance workstation configured with Intel (R) Xeon (R) Silver 4214R (2.40GHz) and Nvidia GeForce RTX 3090 (24GB). 
\subsection{Multi-sensor joint calibration experiments}
\subsubsection{Image distortion correction}
\indent In this paper, a JGS-2 quartz glass is employed to seal the camera lens. The thickness of this glass is 5 millimeters and its refractive index is 1.458. According to the refraction model, the optimal distance between the camera lens and the inner surface of the glass is calculated to be 0.545 mm. Based on this result, the camera imaging model can be approximated as a SVP. Subsequently, the proposed underwater fast distortion correction algorithm is applied to remove distortion from the images. The tool used for distortion correction is a black and white checkerboard with 204 corner points. To verify its effectiveness, a comparative experiment is conducted with the Zhang's calibration method \citep{zhang2002flexible50}. As shown in \autoref{table1}, the reprojection error of our method is 0.4595 pixels, which is 6.5\% lower than that of Zhang's method. In terms of computational efficiency, our method achieves an average running time of $6.107$ × $10^{-6} s/pixel$, nearly 46 times faster than Zhang's method. This remarkable improvement in efficiency primarily results from the proposed fast distortion correction strategy, which optimizes the parameter estimation process. The internal parameters of the camera and IMU are presented in \autoref{table2}.
\begin{figure}[H]
		\centering
		\includegraphics[width=7cm]{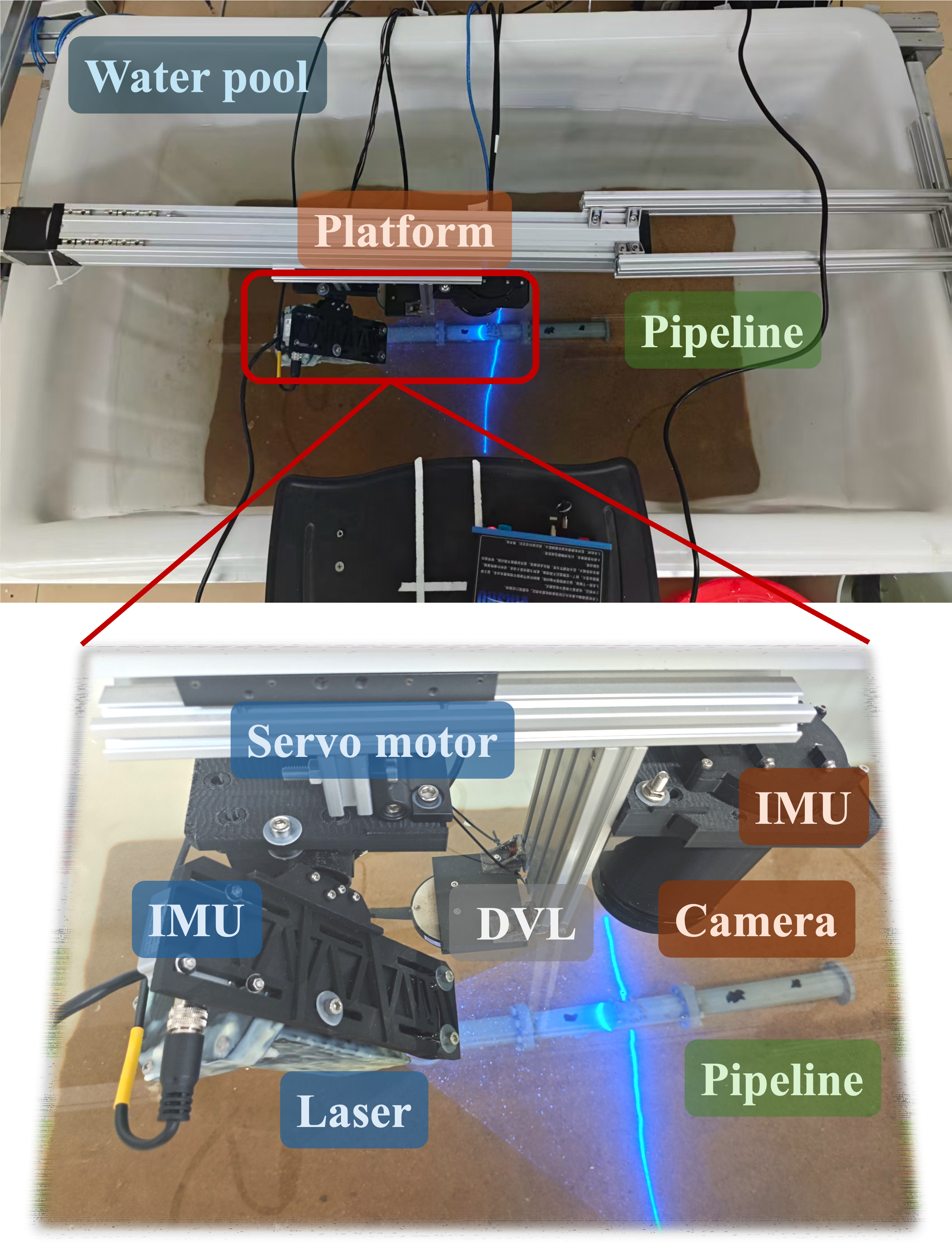}
	\caption{Experimental environment.\label{fig11}}
\end{figure} 
\begin{table}[H]
	\footnotesize
	\caption{Comparative experiments of different image distortion correction methods. Best result is highlighted in bold.\label{tab1}}
	\newcolumntype{C}{>{\centering\arraybackslash}X}
	\begin{tabular}{>{\centering\arraybackslash}m{2cm} >{\centering\arraybackslash}m{2cm}>{\centering\arraybackslash}m{3cm}}
		\toprule[1.5pt]
		\textbf{Method}	& \textbf{Reprojection error/pixel}&\textbf{Average running time/(s/pixel)}\\
		\midrule
		Zhang’s method		& 0.4913 &$2.799$×$10^{-4}$ \\
		Our calibration method	& \textbf{0.4595}& \textbf{$6.107$×$10^{-6}$}\\
		\bottomrule[1.5pt]
	\end{tabular}
	\label{table1}
\end{table}
\subsubsection{Sensors external parameter calibration}
\indent To ensure the accuracy of multi-source information fusion, the external parameters between the camera, IMU, and DVL are calibrated. The transformation matrices obtained from the calibration are summarized in \autoref{table3}. The $\mathbf{T}^{I}_{C}$ represents the transformation from the IMU to the camera coordinate system, and $\mathbf{T}^{C}_{D}$ represents the transformation from the camera to the DVL coordinate system. To verify the calibration accuracy, a comparative experiment is conducted between the proposed method and the physical measurement. A set of structured light 3D imaging data and dead reckoning data is collected for evaluation. The both extrinsic parameters are used to transform the structured light data from the camera coordinate system to the DVL coordinate system. Then, the corresponding position errors are computed. As shown in \autoref{fig13a}, the proposed method achieves lower position errors, indicating superior calibration accuracy compared with the physically measurement. The results in \autoref{fig13b} indicate that the average position error of the proposed method is 0.0394 m, which is 24.4\% lower than the physical measurement. These experimental results demonstrate that our method can significantly enhance the geometric alignment accuracy among multiple sensors.
\begin{table}[H]
	\footnotesize
	\centering
	\caption{Camera and IMU internal parameters.}
	\setlength{\tabcolsep}{0.1mm}{
		\begin{tabular}{>{\centering\arraybackslash}m{2cm} >{\centering\arraybackslash}m{6cm}}
			\toprule[1.5pt]
			\textbf{Internal Parameter} & \textbf{Parameter Matrix} \\
			\midrule
			Camera\_K & 
			$\left[
			\begin{array}{ccc}
				1638.3357 & -0.2092 & 1023.1856 \\
				0 & 1638.1088 & 750.9077 \\
				0 & 0 & 1.0000
			\end{array}
			\right]$ \\
			\midrule
			Camera\_D & 
			$[0.2196,\ 0.2110,\ 0.0066,\ -0.0095,\ 0.9013]$ \\
			\midrule
			IMU & 
			$[0.0017,\ 0.0006,\ 0.0002,\ 0.0000]$ \\
			\bottomrule[1.5pt]
	\end{tabular}}
	\label{table2}
\end{table}
\begin{table}[H]
	\footnotesize
	\centering
	\caption{Calibration results of external parameters.}
	\renewcommand{\arraystretch}{1.6}
	\begin{tabular}{>{\centering\arraybackslash}m{1.5cm} >{\centering\arraybackslash}m{5.3cm}}
		\toprule[1.5pt]
		\textbf{External parameter} & \textbf{Transformation Matrix} \\
		\midrule
		
		$\mathbf{T}^{I}_{C}$ &
		$\left[
		\begin{array}{rrrr}
			-0.996 & 0.073 & -0.049& -0.007 \\
			0.071 & -0.997 & 0.026 & -0.002 \\
			0.051 & -0.022& -0.998 & -0.110
		\end{array}
		\right]$ \\
		
		\midrule
		
		$\mathbf{T}^{C}_{D}$ &
		$\left[
		\begin{array}{rrrr}
			0.976& -0.032& -0.217& -0.124\\
			0.025& 0.999 & -0.037& -0.015\\
			-0.218& -0.030& -0.976& 0.116
		\end{array}
		\right]$ \\
		
		\bottomrule[1.5pt]
	\end{tabular}
	\label{table3}
\end{table}

\begin{figure}[H]
	\centering
	\subfigure[\label{fig13a}]{
		\includegraphics[width=3.5cm]{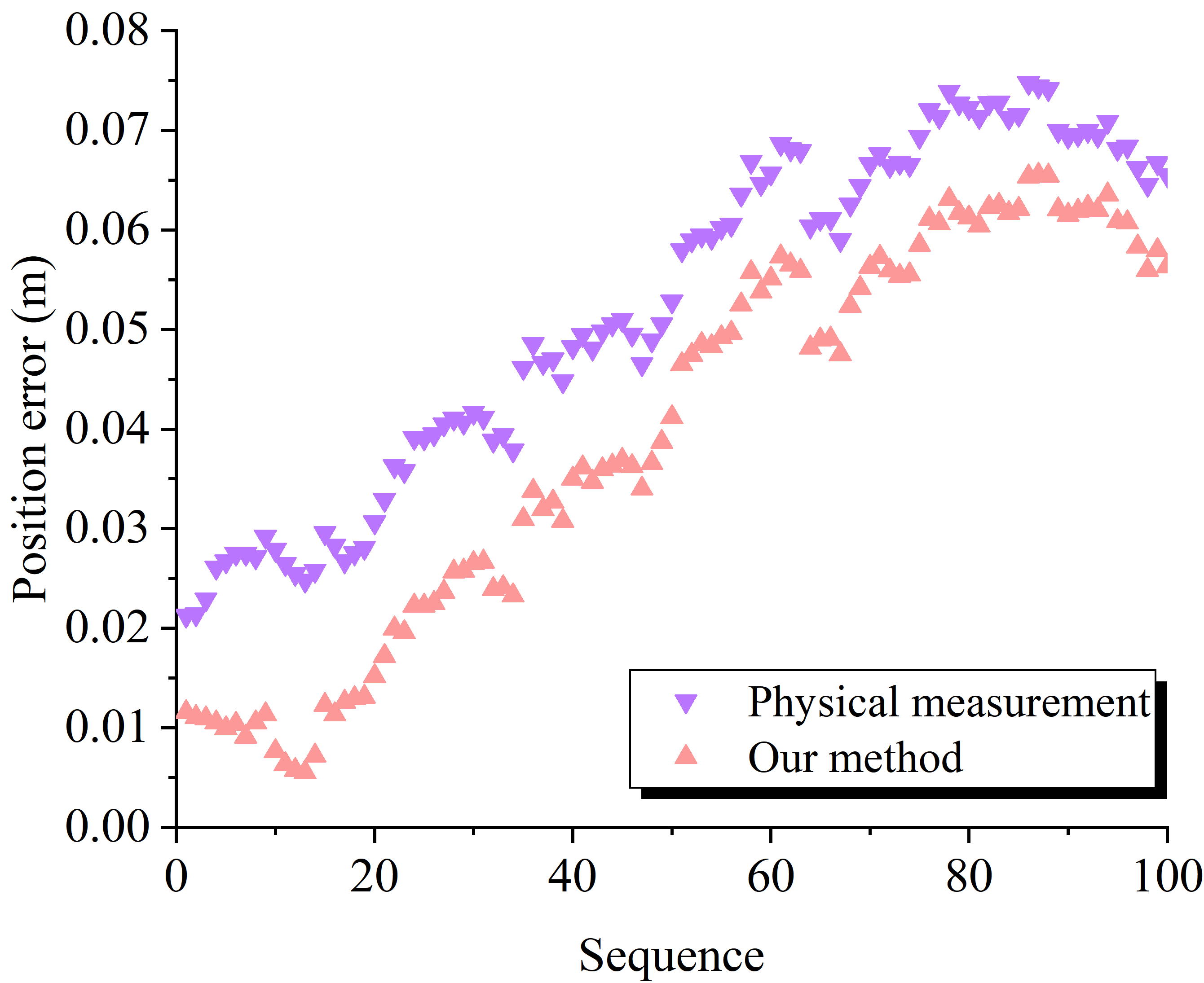}
	}
	\hspace{3mm}
	\subfigure[\label{fig13b}]{
		\includegraphics[width=3.5cm]{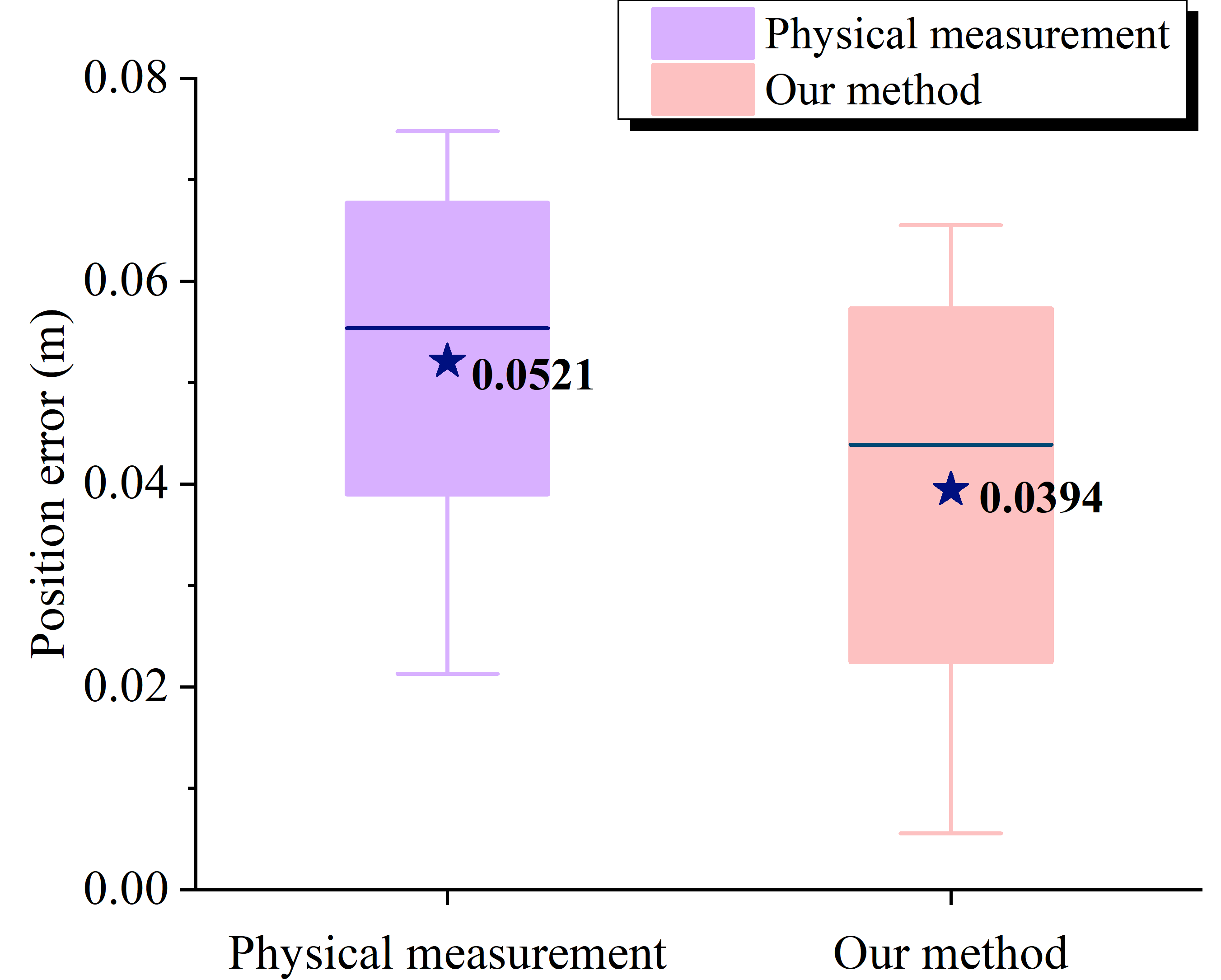}
	}
	\caption{Position errors of different calibration methods. (a) Error distribution. (b) Average error.}
	\label{fig13}
\end{figure}
\subsection{Pose fusion experiments of inertial-acoustic information}
\indent In underwater environments, sensors are often affected by intrinsic drift and unpredictable external interferences. For attitude and position fusion, it is essential to ensure adaptive anti-interference capability. Therefore, this paper proposes a pose fusion method based on AEKF. To validate the effectiveness of the proposed method, comparative experiments are conducted using three methods. During these experiments, the DVL is disabled for a random period to simulate external environmental interference. As illustrated in \autoref{fig14a}-\autoref{fig14c}, the proposed method maintains smooth and stable position estimates along all axes, even during the DVL outage period. The results demonstrate that the proposed method can effectively suppresses drift. In contrast, both DR and EKF exhibit evident oscillations due to the loss of adaptive  constraints. The adaptive method can dynamically adjust the error covariance matrix based on real-time residual statistics. Therefore, it can suppress divergence and enhance system robustness under uncertain measurement conditions. \autoref{table4}-\autoref{table6} summarize the quantitative comparison results of the three methods along the X, Y, and Z axes, including the mean absolute error (MAE), maximum error (ME), standard deviation (SD), and root mean square error (RMSE). In the X-axis, the proposed method achieves an MAE of 0.0455 m, representing reductions of 4.4\% and 10.9\% compared with DR and EKF, respectively. In the Y-axis, the MAE decreases to 0.0147 m, yielding reductions of 93.4\% and 78.3\% relative to DR and EKF. Along the Z-axis, the MAE drops to 0.0304 m, corresponding to improvements of 47.4\% over DR and 37.3\% over EKF. Overall, the AEKF method can achieve the lowest MAE, ME, SD, and RMSE compared with DR and EKF. These results demonstrate that the proposed adaptive fusion framework effectively mitigates inertial drift under unstable acoustic conditions, thereby significantly improving the accuracy and stability of underwater positioning.\vspace{-3mm}
\begin{figure*}[ht]
	\centering
	\subfigure[ \label{fig14a}]{
		\includegraphics[width=5cm]{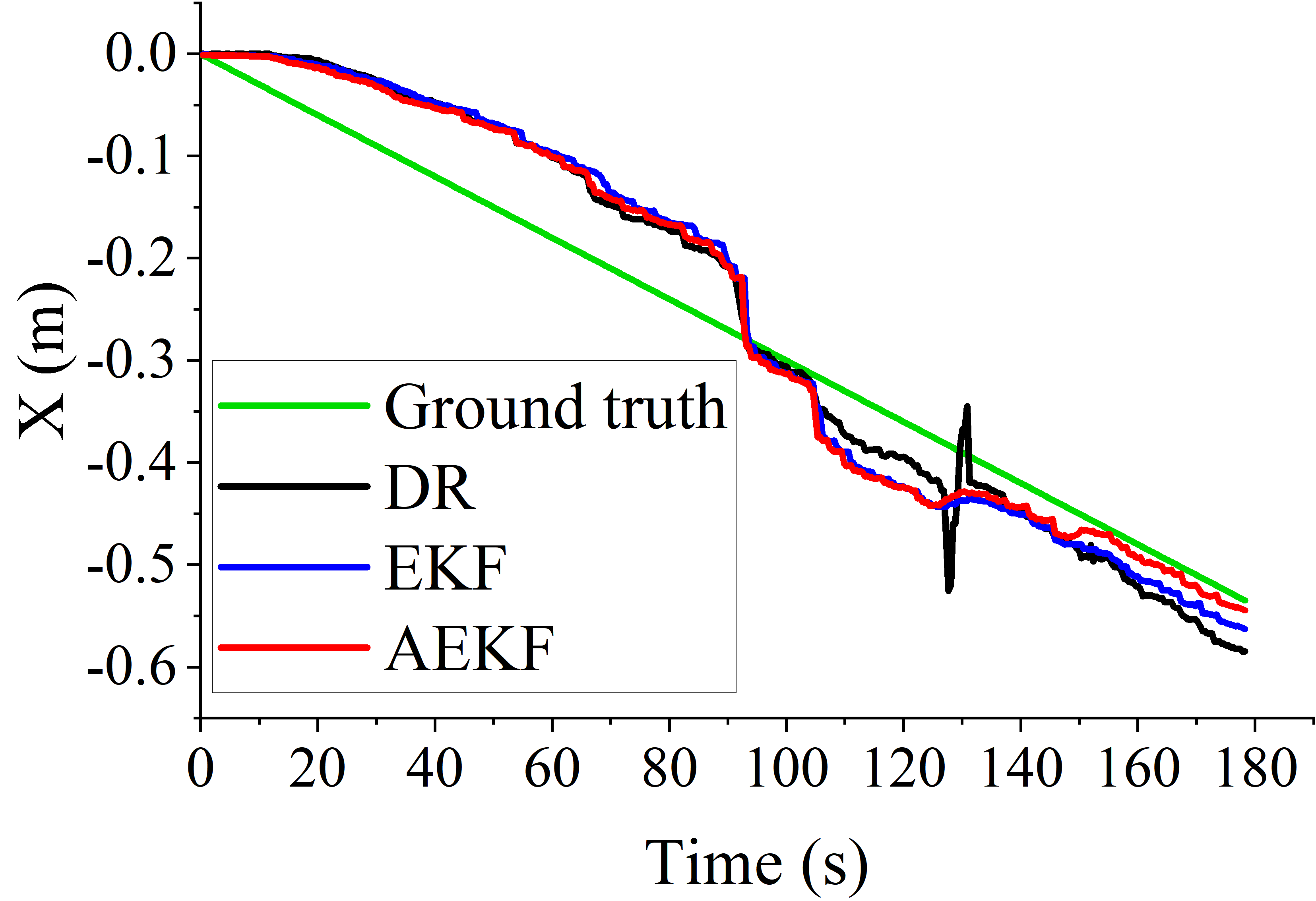}
	}
	\subfigure[\label{fig14b}]{
		\includegraphics[width=5cm]{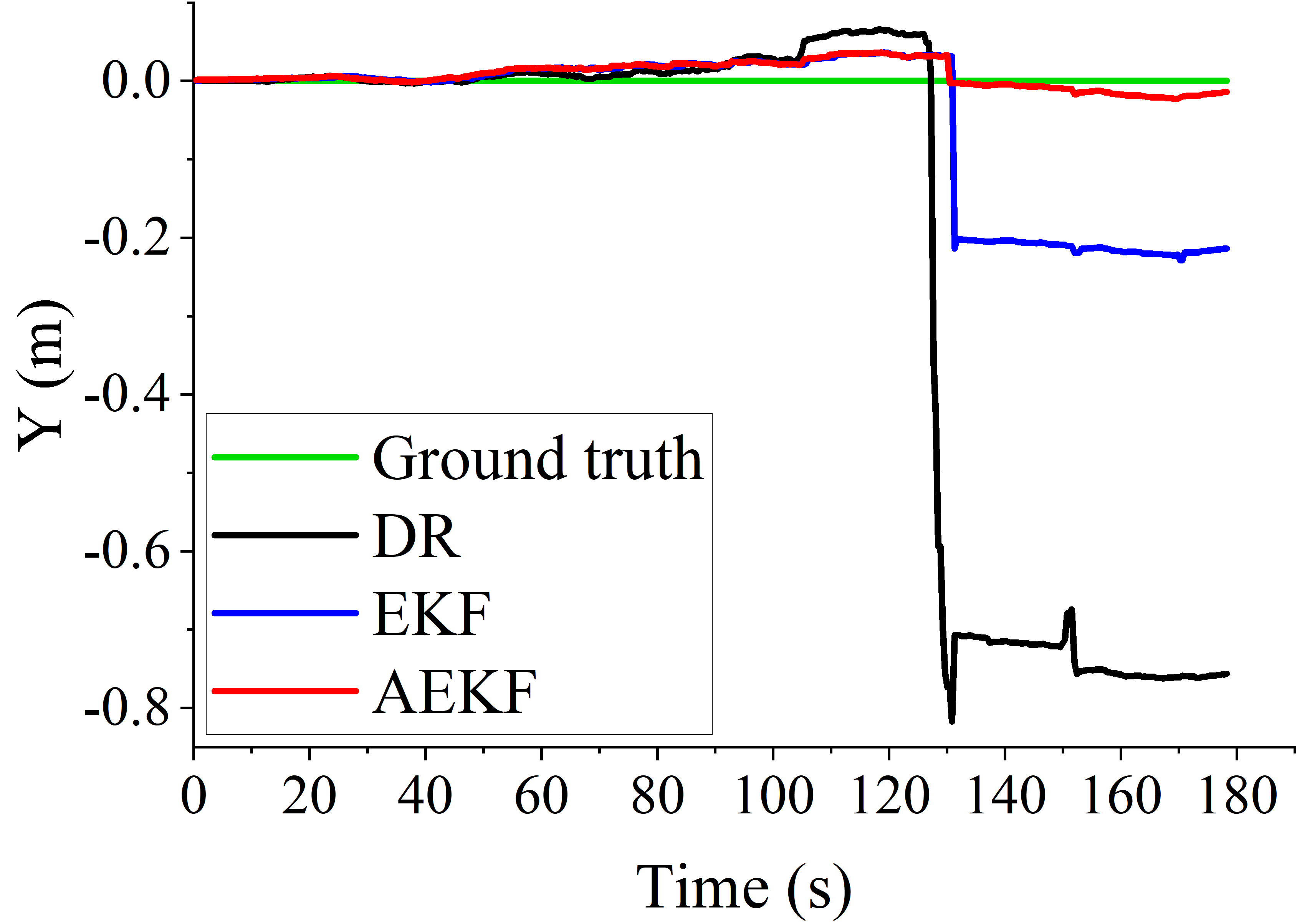}
	}
	\subfigure[\label{fig14c}]{
		\includegraphics[width=5cm]{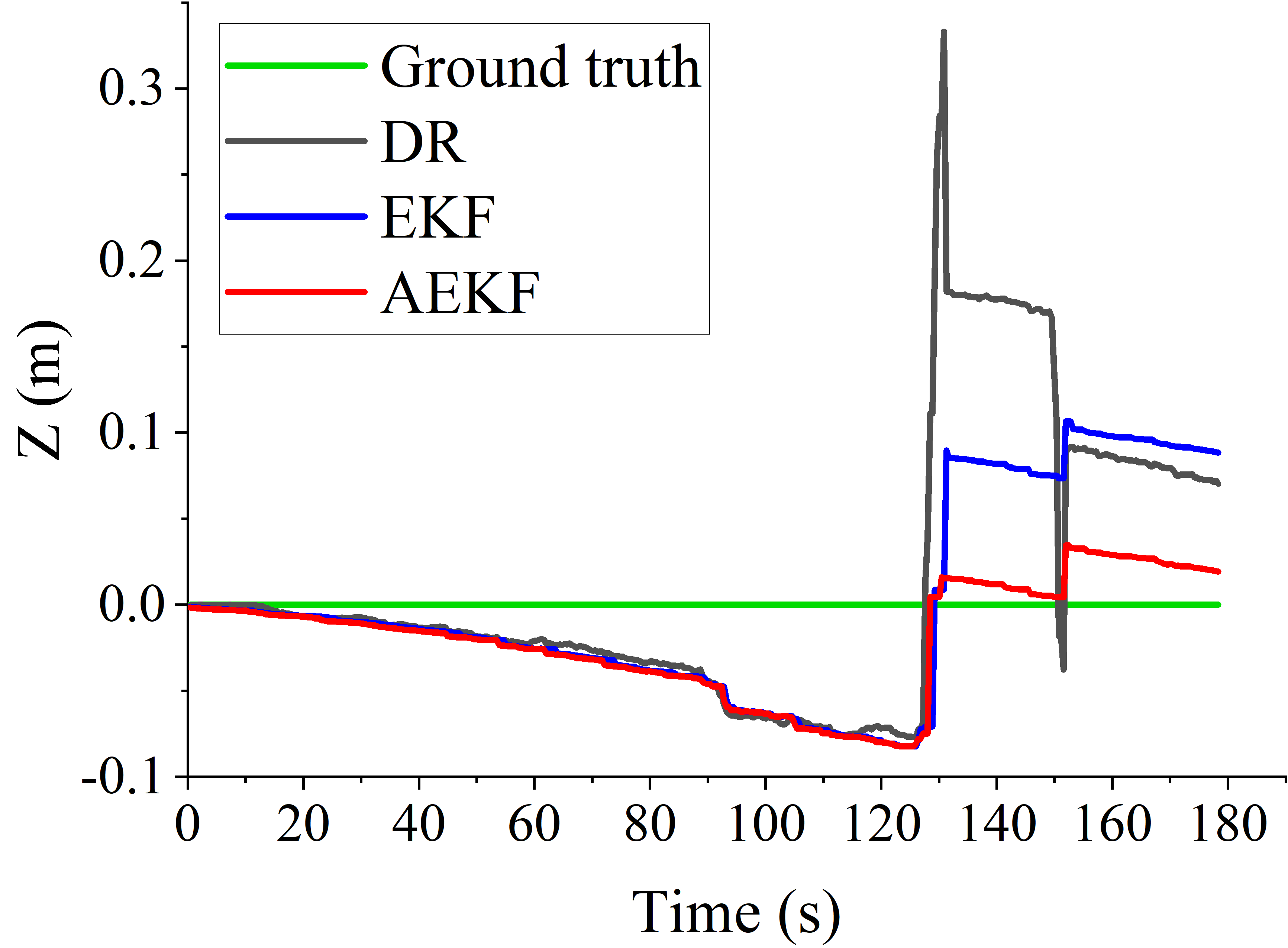}
	}
	\caption{Comparative experiments of different positioning methods. (a) X-axis comparative experiments. (b) Y-axis comparative experiments. (c) Z-axis comparative experiments.}
	\label{fig14}
\end{figure*}

\begin{table}[H]
	\footnotesize
	\centering
	\caption{Comparative experiments of different positioning methods in X-axis. Best result is highlighted in bold.\label{table4}}
	\begin{tabular}{>{\centering\arraybackslash}m{1.5cm}
			>{\centering\arraybackslash}m{1cm}
			>{\centering\arraybackslash}m{1cm}
			>{\centering\arraybackslash}m{1cm}
			>{\centering\arraybackslash}m{1.5cm}}
		\toprule[1.5pt]
		\textbf{Method} & \textbf{MAE/m} & \textbf{ME/m} & \textbf{SD/m} & \textbf{RMSE/m} \\
		\midrule
		DR          & 0.0476 & 0.1426 & 0.0385 & 0.0526 \\
		EKF         & 0.0512 & 0.0869 & 0.0387 & 0.0566 \\
		Our method  & \textbf{0.0455} & \textbf{0.0833} & \textbf{0.0362} & \textbf{0.0523} \\
		\bottomrule[1.5pt]
	\end{tabular}
\end{table}
\vspace{-4mm}
\begin{table}[H] 
	\footnotesize
\centering
\caption{Comparative experiments of different positioning methods in Y-axis. Best result is highlighted in bold.\label{table5}}
\begin{tabular}{>{\centering\arraybackslash}m{1.5cm}
		>{\centering\arraybackslash}m{1cm}
		>{\centering\arraybackslash}m{1cm}
		>{\centering\arraybackslash}m{1cm}
		>{\centering\arraybackslash}m{1.5cm}}
		\toprule[1.5pt]
		\textbf{Method}&
		\textbf{MAE/m}	& 
		\textbf{ME/m}&  \textbf{SD/m}&  \textbf{RMSE/m}\\
		\midrule
		DR	& 0.2217 &0.8173& 0.1148& 0.3930\\
		EKF	& 0.0677&0.2289& 0.0102& 0.0023\\
		Our method & \textbf{0.0147}&\textbf{0.0365}&\textbf{0.0003}&\textbf{0.0004}\\
		\bottomrule[1.5pt]
	\end{tabular}
\end{table}
\vspace{-3mm}
\begin{table}[H] 
	\footnotesize
	\centering
	\caption{Comparative experiments of different positioning methods in Z-axis. Best result is highlighted in bold.\label{table6}}
	\begin{tabular}{>{\centering\arraybackslash}m{1.5cm}
			>{\centering\arraybackslash}m{1cm}
			>{\centering\arraybackslash}m{1cm}
			>{\centering\arraybackslash}m{1cm}
			>{\centering\arraybackslash}m{1.5cm}}
		\toprule[1.5pt]
		\textbf{Method}&
		\textbf{MAE/m}	& 
		\textbf{ME/m}&  \textbf{SD/m}&  \textbf{RMSE/m}\\
		\midrule
		DR	& 0.0578 &0.3334& 0.0063& 0.0804\\
		EKF	& 0.0485&0.1065&0.0035&0.0589\\
		Our method & \textbf{0.0304}&\textbf{0.0823}&\textbf{0.0011}&\textbf{0.0386}\\
		\bottomrule[1.5pt]
	\end{tabular}
\end{table}
\subsection{Underwater 3D imaging accuracy measurement experiments }
To verify the 3D imaging accuracy of the proposed UW-SLD system, three modes of comparative experiments are conducted between our method with Xu’s method~\citep{JQRR202205006_51}, as summarized in \autoref{table9}. During theses experiments, the only change is that Xu's light plane calibration method is adopted to replace our Algorithm \autoref{alg:estimation}, while all other conditions remained unchanged. \autoref{table9} lists the measurement values, absolute errors and relative errors of translation, rotation and translation-rotation imaging in the X, Y and Z directions from top to bottom in sequence. The results demonstrate that our proposed method achieved lower errors than Xu’s method in all modes. In the translation imaging mode, the maximum relative error along the X-axis is 1.37\%, approximately 61.4\% lower than the Xu’s method. In the Y-axis, the maximum relative error is 1.12\%, approximately 64.6\% lower than the Xu’s method. The most pronounced improvement in the Z-axis, where the maximum relative error is reduced from 9.40\% to 2.68\%. In the rotation imaging mode, the advantages of our method become more evident. The maximum relative errors in the X , Y , and Z directions are 2.20\%, 1.76\%, and 1.72\%. These correspond to error reduction rates of 63.0\%, 68.3\%, and 82.7\%. This demonstrates that the proposed method possesses strong robustness to geometric changes during rotational motion. In the translation-rotation imaging mode, the proposed method continues to deliver high accuracy, with maximum relative errors of 1.25\%, 2.90\%, and 1.64\% in the X, Y, and Z directions, respectively. The corresponding reductions are 84.8\%, 55.4\%, and 91.1\%. These verify the adaptability of the proposed method under different imaging trajectories.
\indent According to the all experiments, it is evident that the measurement error of the proposed method remains below 3\% in all cases.   Although the translational–rotational mode introduces more complex geometric relationships and light path variations, the measurement accuracy shows no significant degradation compared with single mode. The visualization imaging results of our method in three modes are shown in \autoref{fig15}.
\vspace{-4mm}
\begin{figure}[H]
		\centering
		\includegraphics[width=8cm]{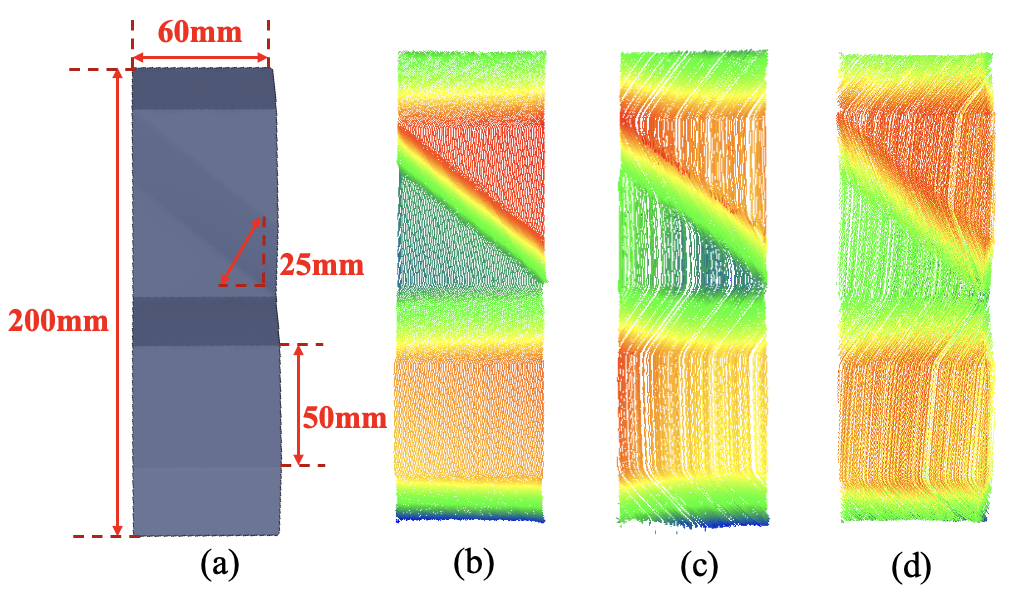}
	\caption{Visualization of imaging results. (a) 3D model. (b) Translation imaging. (c) Rotation imaging. (d) Translation-Rotation imaging. \label{fig15}}
\end{figure} 
\begin{table*}[b!]
	\centering
	\caption{Comparison of detection accuracy under multi-mode 3D imaging. Best result is highlighted in bold.}
	\resizebox{\textwidth}{!}{
		\begin{tabular}{ccccccccc}
			\toprule[1.5pt]
			\multicolumn{1}{c}{\multirow{3}[3]{*}{Direction}} &
			\multicolumn{1}{c}{\multirow{3}[3]{*}{Groud truth/mm}} & \multicolumn{3}{c}{Method proposed by \citep{JQRR202205006_51}} &  & \multicolumn{3}{c}{Our method} \\
			\cmidrule{3-5}\cmidrule{7-9}          
			& & \multicolumn{1}{p{5.19em}}{\centering Measured value/mm} & \multicolumn{1}{p{5.19em}}{\centering Absolute error/mm} & \multicolumn{1}{p{5.19em}}{\centering Relative error/\%} & & \multicolumn{1}{p{5.19em}}{\centering Measured value/mm} & \multicolumn{1}{p{5.19em}}{\centering Absolute error/mm} & \multicolumn{1}{p{5.19em}}{\centering Relative error/\%}  \\  
			\cmidrule{1-9}     
			\makecell[t]{$X^t_1$} & \makecell[t]{60} & \makecell[t]{61.49} & \makecell[t]{1.49} & \makecell[t]{2.48} & & \makecell[t]{60.18} & \makecell[t]{0.18} & \makecell[t]{\textbf{0.30}}\\
			\makecell[t]{$X^t_2$} & \makecell[t]{60} & \makecell[t]{61.91} & \makecell[t]{1.91} & \makecell[t]{3.18}  & & \makecell[t]{59.18} & \makecell[t]{0.82} & \makecell[t]{\textbf{1.37}} \\
			\makecell[t]{$X^t_3$} & \makecell[t]{60} & \makecell[t]{62.13} & \makecell[t]{2.13} & \makecell[t]{3.55}  & & \makecell[t]{59.65} & \makecell[t]{0.35} & \makecell[t]{\textbf{0.58}} \\
			\makecell[t]{$Y^t_1$} & \makecell[t]{200} & \makecell[t]{197.69} & \makecell[t]{2.31} & \makecell[t]{1.16}  & & \makecell[t]{199.59} & \makecell[t]{0.41} & \makecell[t]{\textbf{0.21}}\\
			\makecell[t]{$Y^t_2$} & \makecell[t]{200} & \makecell[t]{198.86} & \makecell[t]{1.14} & \makecell[t]{0.57}  & & \makecell[t]{198.97} & \makecell[t]{1.03} & \makecell[t]{\textbf{0.52}}\\
			\makecell[t]{$Y^t_3$} & \makecell[t]{50} & \makecell[t]{48.42} & \makecell[t]{1.58} & \makecell[t]{3.16}  & & \makecell[t]{50.56} & \makecell[t]{0.56} & \makecell[t]{\textbf{1.12}}\\
			\makecell[t]{$Z^t_1$} & \makecell[t]{25} & \makecell[t]{25.77} & \makecell[t]{0.77} & \makecell[t]{3.08}  & & \makecell[t]{25.67} & \makecell[t]{0.67} & \makecell[t]{\textbf{2.68}}\\
			\makecell[t]{$Z^t_2$} & \makecell[t]{25} & \makecell[t]{22.65} & \makecell[t]{2.35} & \makecell[t]{9.40}  & & \makecell[t]{24.54} & \makecell[t]{0.46} & \makecell[t]{\textbf{1.84}}\\
			\makecell[t]{$Z^t_3$} & \makecell[t]{25} & \makecell[t]{25.89} & \makecell[t]{0.89} & \makecell[t]{3.56}  & & \makecell[t]{25.58} & \makecell[t]{0.58} & \makecell[t]{\textbf{2.32}}\\
			\bottomrule[0.5pt]	
			\addlinespace[3pt]
			\makecell[t]{$X^r_1$} & \makecell[t]{60} & \makecell[t]{56.80} & \makecell[t]{3.20} & \makecell[t]{5.33} & & \makecell[t]{58.96} & \makecell[t]{1.04} & \makecell[t]{\textbf{1.73}}\\
			\makecell[t]{$X^r_2$} & \makecell[t]{60} & \makecell[t]{57.29} & \makecell[t]{2.71} & \makecell[t]{4.52}  & & \makecell[t]{59.09} & \makecell[t]{0.91} & \makecell[t]{\textbf{1.52}} \\
			\makecell[t]{$X^r_3$} & \makecell[t]{60} & \makecell[t]{56.43} & \makecell[t]{3.57} & \makecell[t]{5.95}  & & \makecell[t]{58.68} & \makecell[t]{1.32} & \makecell[t]{\textbf{2.20}} \\
			\makecell[t]{$Y^r_1$} & \makecell[t]{200} & \makecell[t]{196.98} & \makecell[t]{3.02} & \makecell[t]{1.51}  & & \makecell[t]{199.68} & \makecell[t]{0.32} & \makecell[t]{\textbf{0.16}}\\
			\makecell[t]{$Y^r_2$} & \makecell[t]{200} & \makecell[t]{198.45} & \makecell[t]{1.55} & \makecell[t]{0.78}  & & \makecell[t]{199.63} & \makecell[t]{0.37} & \makecell[t]{\textbf{0.19}}\\
			\makecell[t]{$Y^r_3$} & \makecell[t]{50} & \makecell[t]{47.22} & \makecell[t]{2.78} & \makecell[t]{5.56}  & & \makecell[t]{49.12} & \makecell[t]{0.88} & \makecell[t]{\textbf{1.76}}\\
			\makecell[t]{$Z^r_1$} & \makecell[t]{25} & \makecell[t]{26.98} & \makecell[t]{1.98} & \makecell[t]{7.92}  & & \makecell[t]{25.35} & \makecell[t]{0.35} & \makecell[t]{\textbf{1.40}}\\
			\makecell[t]{$Z^r_2$} & \makecell[t]{25} & \makecell[t]{23.98} & \makecell[t]{1.02} & \makecell[t]{4.08}  & & \makecell[t]{25.02} & \makecell[t]{0.02} & \makecell[t]{\textbf{0.08}}\\
			\makecell[t]{$Z^r_3$} & \makecell[t]{25} & \makecell[t]{27.48} & \makecell[t]{2.48} & \makecell[t]{9.92}  & & \makecell[t]{24.57} & \makecell[t]{0.43} & \makecell[t]{\textbf{1.72}}\\
			\bottomrule[0.5pt]	
			\addlinespace[3pt]
			\makecell[t]{$X^{tr}_1$} & \makecell[t]{60} & \makecell[t]{56.48} & \makecell[t]{3.52} & \makecell[t]{5.87} & & \makecell[t]{59.94} & \makecell[t]{0.06} & \makecell[t]{\textbf{0.10}}\\
			\makecell[t]{$X^{tr}_2$} & \makecell[t]{60} & \makecell[t]{55.08} & \makecell[t]{4.92} & \makecell[t]{8.20}  & & \makecell[t]{60.68} & \makecell[t]{0.68} & \makecell[t]{\textbf{1.13}} \\
			\makecell[t]{$X^{tr}_3$} & \makecell[t]{60} & \makecell[t]{57.60} & \makecell[t]{2.40} & \makecell[t]{4.00}  & & \makecell[t]{60.75} & \makecell[t]{0.75} & \makecell[t]{\textbf{1.25}} \\
			\makecell[t]{$Y^{tr}_1$} & \makecell[t]{200} & \makecell[t]{197.69} & \makecell[t]{2.31} & \makecell[t]{1.16}  & & \makecell[t]{199.02} & \makecell[t]{0.98} & \makecell[t]{\textbf{0.49}}\\
			\makecell[t]{$Y^{tr}_2$} & \makecell[t]{200} & \makecell[t]{204.39} & \makecell[t]{4.39} & \makecell[t]{2.19}  & & \makecell[t]{198.45} & \makecell[t]{1.55} & \makecell[t]{\textbf{0.78}}\\
			\makecell[t]{$Y^{tr}_3$} & \makecell[t]{50} & \makecell[t]{46.75} & \makecell[t]{3.25} & \makecell[t]{6.50}  & & \makecell[t]{48.55} & \makecell[t]{1.45} & \makecell[t]{\textbf{2.90}}\\
			\makecell[t]{$Z^{tr}_1$} & \makecell[t]{25} & \makecell[t]{29.55} & \makecell[t]{4.55} & \makecell[t]{18.20}  & & \makecell[t]{25.24} & \makecell[t]{0.24} & \makecell[t]{\textbf{0.96}}\\
			\makecell[t]{$Z^{tr}_2$} & \makecell[t]{25} & \makecell[t]{29.59} & \makecell[t]{4.59} & \makecell[t]{18.36}  & & \makecell[t]{25.25} & \makecell[t]{0.25} & \makecell[t]{\textbf{1.00}}\\
			\makecell[t]{$Z^{tr}_3$} & \makecell[t]{25} & \makecell[t]{29.06} & \makecell[t]{4.06} & \makecell[t]{16.24}  & & \makecell[t]{24.59} & \makecell[t]{0.41} & \makecell[t]{\textbf{1.64}}\\
			\bottomrule[1.5pt]	
		\end{tabular}%
	}
	\label{table9}%
\end{table*}%
\subsection{3D imaging experiments for underwater pipeline detection}
\indent To further verify the applicability of the UW-SLD system in real underwater scenarios, experiments are conducted on multiple types of pipelines. These experimental pipelines include small-diameter pipelines with flange connections (Pipelines A and B) and large-diameter pipelines with surface anomalies (Leakage A–B, Attachment A). These different surface anomalies include leaks, depressions, and attachments, as illustrated in \autoref{fig16}. Each pipeline is detected using two modes: translation and translation–rotation imaging. The quantitative evaluations of pipeline detection are performed, as summarized in \autoref{table10}. For Pipeline A and Pipeline B with multiple flange connections, the measured total lengths under the translation mode are 500.31 mm and 499.31 mm, respectively. The corresponding relative errors are 0.06\% and 0.14\%, respectively. In the translation–rotation mode, the total lengths are 501.22 mm and 499.55 mm. The relative errors slightly changed to 0.24\% and 0.11\%. This indicates that the accuracy change caused by rotational motion is very small. For the large-diameter pipelines, the system also achieved high accuracy. The relative errors in the translation mode are 0.29\%, 0.06\%, and 0.01\%, while in the translation–rotation mode they are 0.23\%, 0.16\%, and 0.25\%, respectively. The minor variations between the two imaging modes demonstrate that the system maintains consistent performance regardless of pipeline diameter or surface morphology. 
\begin{table*}[b!]
	\centering
	\caption{Detection results of different pipelines under various imaging modes. Best result is highlighted in bold.}
	\resizebox{\textwidth}{!}{
		\begin{tabular}{ccccccccccc}
			\toprule[1.5pt]
			\multirow{3}{*}{Class} & \multirow{3}{*}{Number} &
			\multirow{3}{*}{Groud truth/mm} & \multicolumn{2}{c}{Measurement value/mm} &  & \multicolumn{2}{c}{Absolute error/mm} &  & \multicolumn{2}{c}{Relative error/\%}\\
			\cmidrule{4-5}\cmidrule{7-8}        \cmidrule{10-11}  
			& & & T & T-R & & T & T-R & & T & T-R\\  
			\cmidrule{1-11}     
			Pipeline A & $1$ & 160 & 160.47 & 160.58 & & 0.47 & 0.58& & \textbf{0.29}&0.36 \\
			$ $ & $2$ & 160 & 160.26 & 159.94 & & 0.26 & 0.06 & & 0.16&\textbf{0.04}\\
			$ $ & $3$ & 160 & 159.93 & 160.09 & &0.07 & 0.09 & & \textbf{0.04}&0.06\\
			$ $ & $4$ & 10 & 9.74 & 9.96& & 0.26 & 0.04& & 2.60& \textbf{0.40}\\
			$ $ & $5$ & 10 & 9.91 & 10.65 & & 0.09 & 0.65 & &\textbf{0.90} &6.50\\
			$ $ & total & 500 & 500.31 & 501.22 & & 0.31 & 1.22& & \textbf{0.06}&0.24 \\
			Pipeline B & $1$ & 160 & 160.19 & 159.44 & & 0.19& 0.56 & & \textbf{0.12}&0.35\\
			$ $ & $2$ & 160 & 159.58 & 160.19& & 0.42& 0.19 & & 0.26&\textbf{0.12}\\
			$ $ & $3$ & 160 & 159.64 &159.60 & & 0.36 & 0.40 & & \textbf{0.23}&0.25\\
			$ $ & $4$ & 10 & 9.89 & 10.59& & 0.11 & 0.59& &\textbf{1.10}&5.90 \\
			$ $ & $5$ & 10 & 10.01 & 9.73& & 0.01& 0.27 & & \textbf{0.10}&2.7\\
			$ $ & total & 500 & 499.31 & 499.55 & & 0.69 & 0.45 & & 0.14&\textbf{0.11}\\
			Leakage A & $1$ & 305 & 304.13 & 305.69 & & 0.87 & 0.69 & & 0.29&\textbf{0.23}\\
			Leakage B & $1$ & 305 & 305.17 & 305.48 & & 0.17 & 0.48& & \textbf{0.06}&0.16 \\
			Attachment A & $1$ & 452 & 451.97 & 453.14 & & 0.03 & 1.14& & \textbf{0.01}&0.25 \\
			
			\bottomrule[1.5pt]	
		\end{tabular}%
	}
	\label{table10}%
\end{table*}%
Abnormalities in underwater pipelines are manifested not only as surface defects but also as abnormal spatial relationships with the surrounding ground surface. Typically, underwater pipelines may exhibit buried, exposed, or uplifted states, each indicating potential structural or safety risks. As illustrated in \autoref{fig16} (a) and (b), the depth distribution shows that segment 1 is nearly level with the surrounding terrain over a broad area, indicating a certain degree of burial. Segment 2 maintains a consistent elevation relative to its local environment and is therefore considered normal. In contrast, segment 3 presents a pronounced height difference from the nearby ground surface, suggesting an exposed condition. From the \autoref{fig16} (f) and (g), an uplifted state can be observed in segment B2, while B1 is buried and B3 appears normal. The noticeable increase in elevation difference around B2 indicates that the pipeline has lifted relative to its adjacent sections. 
\begin{figure*}[ht]
		\centering
		\includegraphics[width=16cm]{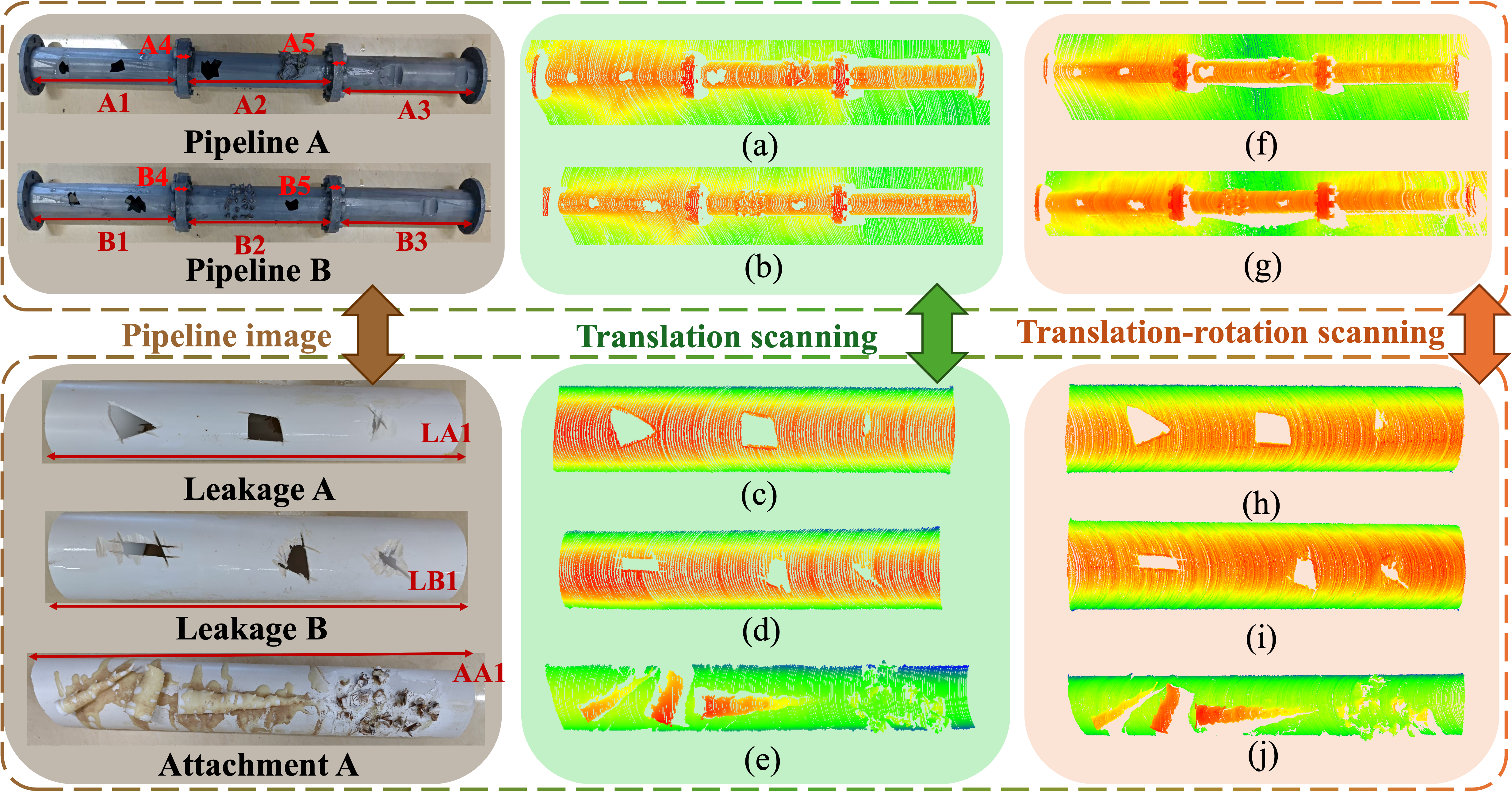}
	\caption{Visualization of different pipelines under various 3D imaging modes. (a)-(e) Translation imaging results. (f)-(j) Translation-Rotation imaging results. \label{fig16}}
\end{figure*} 
\indent To evaluate the environmental adaptability of the UW-SLD system, different pipeline experiments are performed under different water depths. As shown in \autoref{fig17}, the tested objects include multiple categories of pipelines with different surface characteristics. The system is used to perform complete 3D imaging of these pipelines at two depths of 0.36-0.39m and 0.50-0.55m. The quantitative results are summarized in \autoref{table11}. From the \autoref{table11}, it can be observed that the UW-SLD system maintains high measurement accuracy in both shallow and deep water. For the leaky pipelines, the relative error in shallow water is 0.23\% and 0.16\%, while in deep water it slightly increases to 0.38\% and 0.19\%. For the pipeline with attachments, the relative error increases from 0.25\% in shallow water to 0.32\% in deep water. The above experiment indicates that more disturbances are introduced due to the increase in water depth. For curved pipelines with more complex surface geometries, the measurement accuracy remains consistently high in both environments. For Pipeline A with multiple flange connections, the total relative errors in shallow and deep water are 0.24\% and 0.66\%, respectively. This slight variation in error further demonstrates the robustness of the system against refraction and reflection interference under deeper water conditions.
\begin{figure}[H]
		\centering
		\includegraphics[width=8cm]{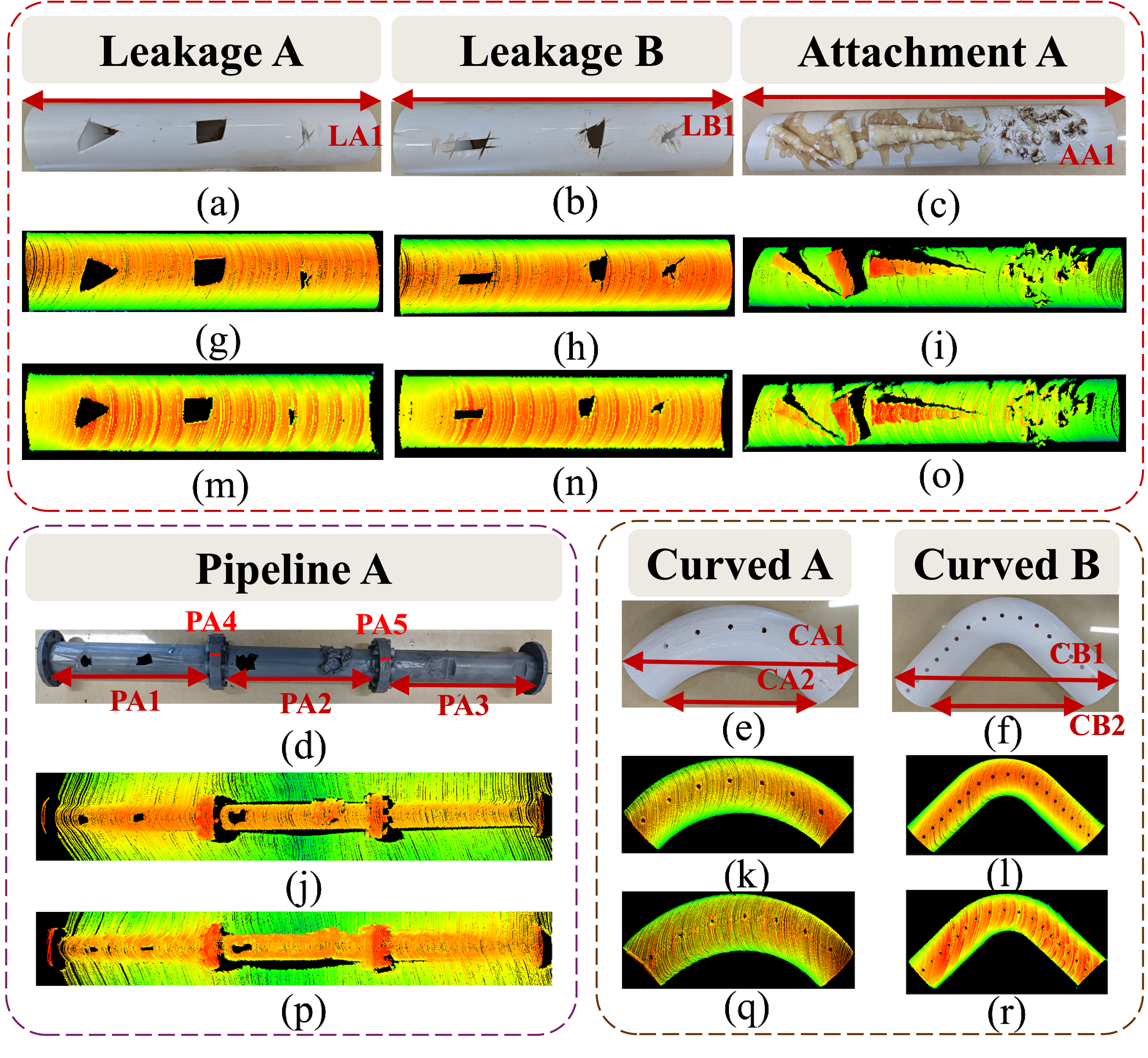}
	\caption{Visualization of different pipelines at various water depths. (a)-(f) Original image. (g)-(l) Shallow water imaging. (m)-(r) Deep water imaging. \label{fig17}}
\end{figure}
\begin{table*}[ht]
	\centering
	\caption{3D imaging results of different pipelines at different water depths. Best result is highlighted in bold.}
	\resizebox{\textwidth}{!}{
		\begin{tabular}{ccccccccccc}
			\toprule[1.5pt]
			\multirow{3}{*}{Class} & \multirow{3}{*}{Number} &
			\multirow{3}{*}{Groud truth/mm} & \multicolumn{2}{c}{Distance/mm} &  &  \multicolumn{2}{c}{Absolute error/mm} & & 
			\multicolumn{2}{c}{Relative error/\%}\\
			\cmidrule{4-5}\cmidrule{7-8}        \cmidrule{10-11}          
			& & & Shallow & Deep & & Shallow & Deep & & Shallow & Deep\\  
			\cmidrule{1-11}     
			Leakage A & $1$ & 305 & 305.69 & 306.16 & & 0.69 & 1.16 &&\textbf{0.23} &0.38 \\
			Leakage B & $1$ & 305 & 305.48 & 305.60 & & 0.48 & 0.60 &&\textbf{0.16} &0.19\\
			Attachment A & $1$ & 452 & 453.14 & 453.44 & & 1.14 & 1.44 &&\textbf{0.25}& 0.32  \\
			Curved A & $1$ & 200 & 200.21 & 199.47 & & 0.21 & 0.53 &&\textbf{0.11}&0.27  \\
			$ $  & $2$ & 280 & 280.61 & 279.64 & & 0.61 & 0.36 && 0.22 &\textbf{0.13} \\
			Curved B & $1$ & 243 & 243.22 & 242.73 & & 0.22 & 0.27 && \textbf{0.09}&0.11  \\
			$ $ & $2$ & 335 & 334.74 & 334.89 & & 0.26 & 0.11 && 0.08&\textbf{0.03}\\
			Pipeline A & $1$ & 160 & 160.58 & 160.86 & & 0.58 & 0.86 & &\textbf{0.36} & 0.54 \\
			$ $ & $2$ & 160 & 159.94 &161.03 & & 0.06 & 1.03& &\textbf{0.04} &0.64 \\
			$ $ & $3$ & 160 & 160.09 & 160.54 & & 0.09 & 0.54 & &\textbf{0.06} &0.34 \\
			$ $ & $4$ & 10 & 9.96 & 10.74 & & 0.04 & 0.74 & &\textbf{0.40} &7.40 \\
			$ $ & $5$ & 10 & 10.65 & 10.14 & & 0.65 & 0.14 & &6.50 &\textbf{1.40} \\
			$ $ & total & 500 & 501.22 & 503.31 & & 1.22 & 3.31 & &\textbf{0.24} &0.66 \\
			
			\bottomrule[1.5pt]	
		\end{tabular}%
	}
	\label{table11}%
\end{table*}%
\subsection{Underwater pipeline edge detection experiments}
\indent As shown in \autoref{fig18}, the image is processed by the edge detection network. To clearly show the effectiveness of this network, image enhancement is introduced on the original image. The \autoref{fig18} (a)-(c) are the original images, while (d)-(f) are the enhanced images. The results indicate that the extracted edge curves can be aligned clearly with the actual pipeline contour. Even if changes in observation angle or light scattering, the network model has strong robustness. The quantitative evaluation results are listed in \autoref{table12}. The detection accuracy of this network is 99.04\% with no FP and FN, while maintaining a real-time processing speed of 218.92 frames per second. These results demonstrate that the proposed network can achieve high precision and efficiency in underwater edge detection. The combination of structured light imaging and high-performance neural networks can provide reliable support for intelligent and autonomous detection in multi-task scenarios.
\vspace{-3mm}
\begin{figure}[H]
		\centering
		\includegraphics[width= 7 cm]{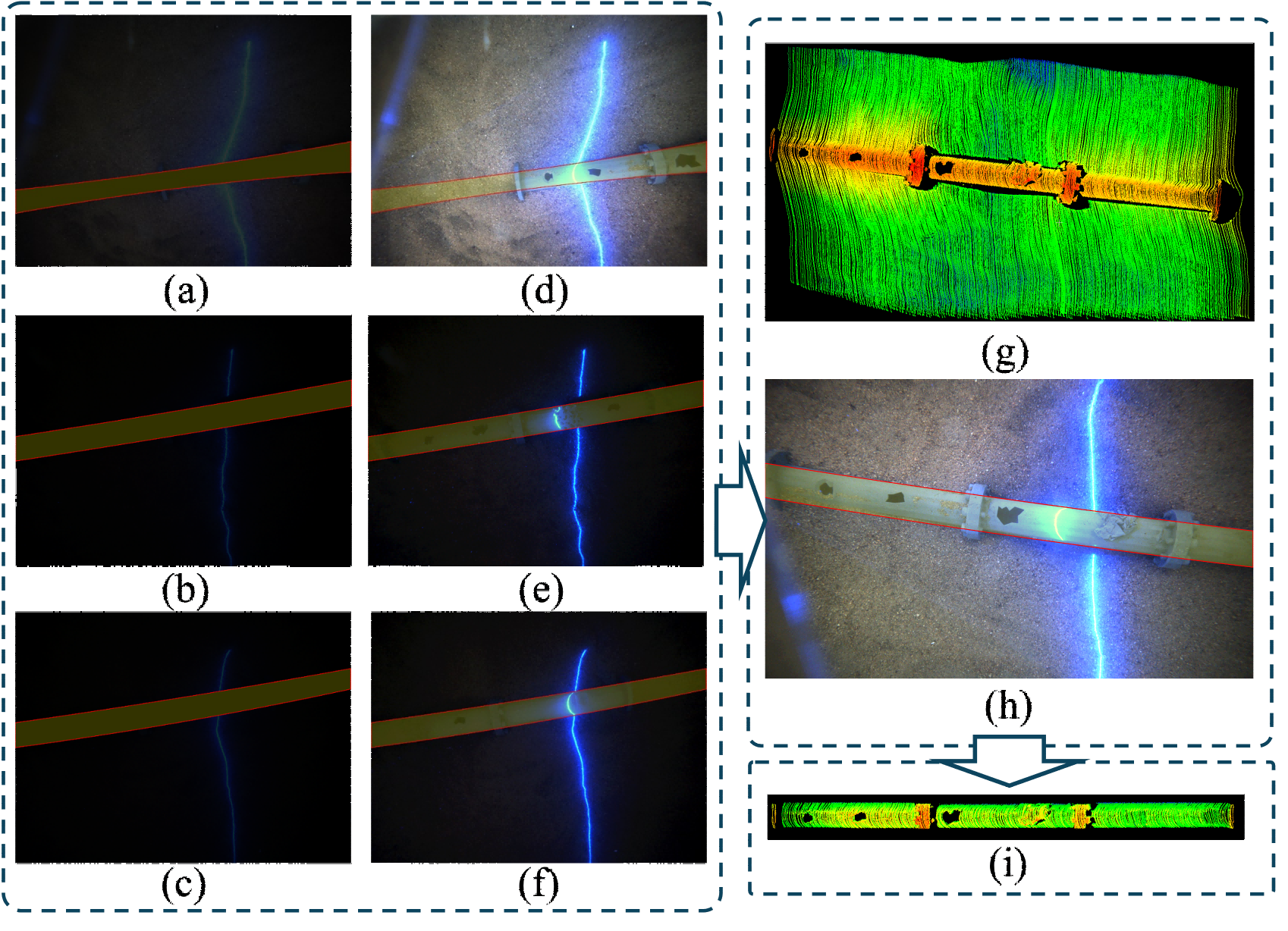}
	\caption{Visualization results of the underwater edge detection network. (a) and (d) Edge detection of Pipeline A. (b)-(c) and (e)-(f) Edege detection of Pipeline B. (g)-(i) 3D point cloud extraction flowchart for the target pipeline using edge detection.\label{fig18}}
\end{figure} 
\vspace{-5mm}
\begin{table}[H]
	\footnotesize
	\centering
	\caption{Experimental results of the edge detection network.}
	\label{table12}
	\newcolumntype{C}{>{\centering\arraybackslash}X}
	\begin{tabular}{ccccc}
		\toprule[1.5pt]
		Method & Accuracy/\% & FP& FN & FPS /f.s$^{-1}$ \\
		\midrule
		Our network&99.04  & 0.00 & 0.00 & 218.92\\
		
		\bottomrule[1.5pt]
	\end{tabular}
\end{table}
\vspace{-5mm}
\begin{table}[H]
	\footnotesize
	\centering
	\caption{Comparision of multi-frame registration results under dynamic motion. Best result is highlighted in bold.}
	\label{table13}
	\newcolumntype{C}{>{\centering\arraybackslash}X}
	\begin{tabular}{ccccc@{\hskip 12pt}ccc}
		\toprule[1.5pt]
		\multirow{2}{*}{} 
		& \multicolumn{3}{c}{Pipeline A} 
		& \multicolumn{3}{c}{Pipeline B} \\
		\cmidrule{2-4} \cmidrule{5-7}
		& ICP & GICP & Ours
		& ICP & GICP & Ours \\
		\midrule
		RMSE  & 7.84 & 7.02 & \textbf{1.08} & 7.21& 6.24& \textbf{2.31}\\
		\bottomrule[1.5pt]
	\end{tabular}
\end{table}
\vspace{-5mm}
\begin{table}[H]
	\footnotesize
	\centering
	\caption{Comparision of overall registration results under dynamic motion. Best result is highlighted in bold.}
	\label{table14}
	\newcolumntype{C}{>{\centering\arraybackslash}X}
	\begin{tabular}{cccccc}
		\toprule[1.5pt]
		& Only DR & AEKF& ICP & GICP & Ours \\
		\midrule
		RMSE (A)&1.28  & 1.27 & 1.69 & 1.63 & \textbf{1.03} \\
		RMSE (B)  &1.23  & 0.94 & 0.78 & 0.69 & \textbf{0.69} \\
		\bottomrule[1.5pt]
	\end{tabular}
\end{table}
\begin{figure}[H]
	\centering
		\includegraphics[width=7.5cm]{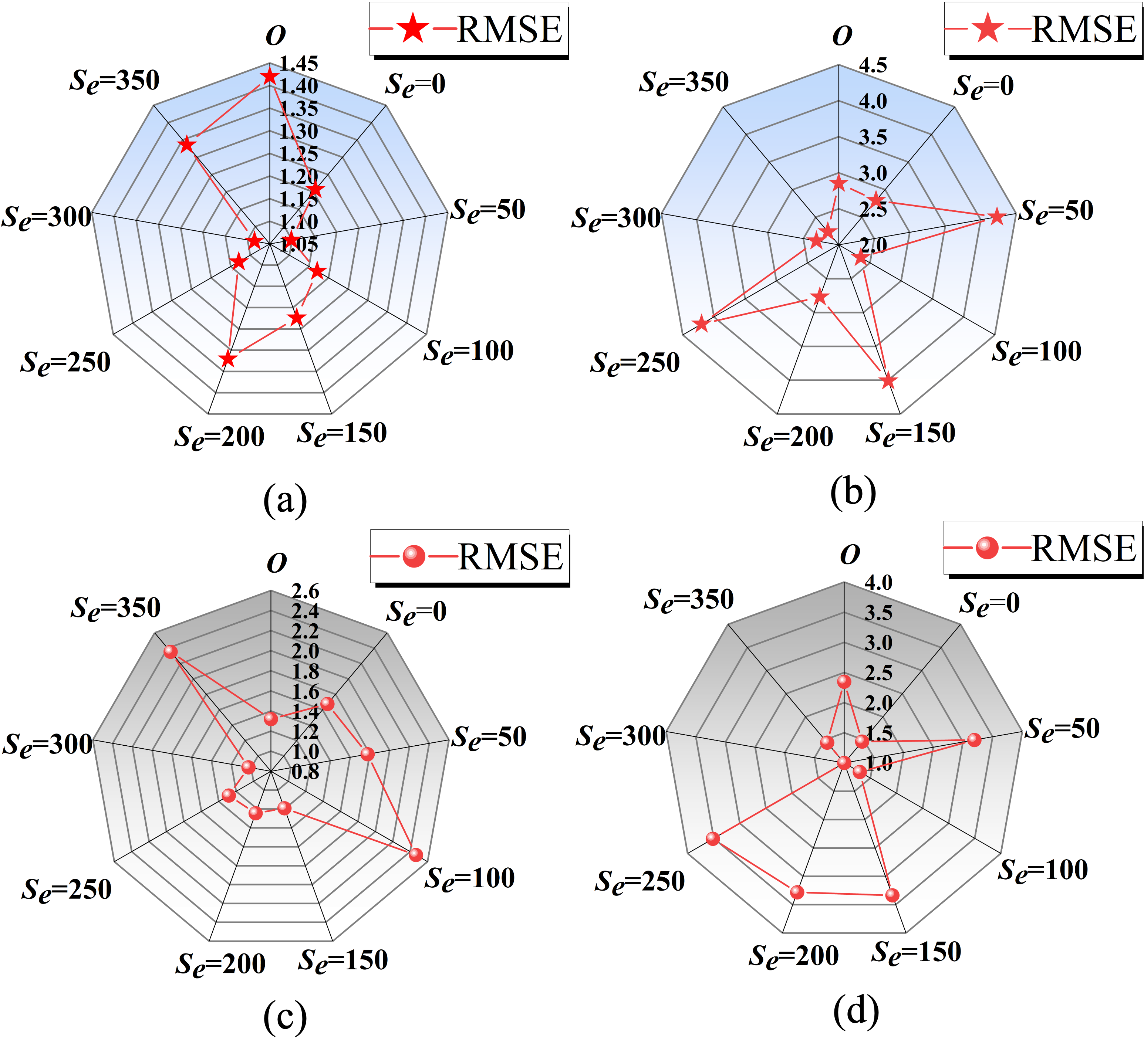}

	\caption{Comparison of registration results under different thresholds. (a) and (b) Frame-by-frame registration results of Pipeline A and B. (c) and (d) Overall registration results of Pipeline A and B. }
	\label{fig19}
\end{figure}
\vspace{-5mm}
\subsection{Underwater pipeline point cloud registration experiments}
\indent In variable-speed imaging, the setting of the pixel-level distance threshold after edge detection directly determines the size of the point set used for matching. A properly selected threshold is expected to maximize the preservation of structural features while suppressing external interference, thereby improving the input quality for the registration algorithm. First, the distance threshold $\epsilon$ is optimized through a grid search to obtain the value that minimizes the registration error under the current conditions. Then, the performances of ICP, GICP, and the proposed method are compared using the same setup. The results of the threshold search are shown in \autoref{fig19}. \autoref{fig19} (a) and \autoref{fig19} (b) show the frame-by-frame registration RMSE, while \autoref{fig19} (c) and \autoref{fig19} (d) present the global registration RMSE. The reference point cloud in the overall registration is obtained through uniform imaging. It should be noted that in \autoref{fig19} (b), the frame-wise RMSE at $\epsilon=300$ is slightly higher than that at 350 pixels. However, superior overall registration performance is achieved at 300 pixels, as shown in \autoref{fig19} (d). By integrating both frame-by-frame and overall registration outcomes, $\epsilon=300$ is selected as the unified threshold, as it best reflects the comprehensive performance of the system under variable-speed conditions.
\begin{figure*}[b!]
	\centering
	\includegraphics[width=16cm]{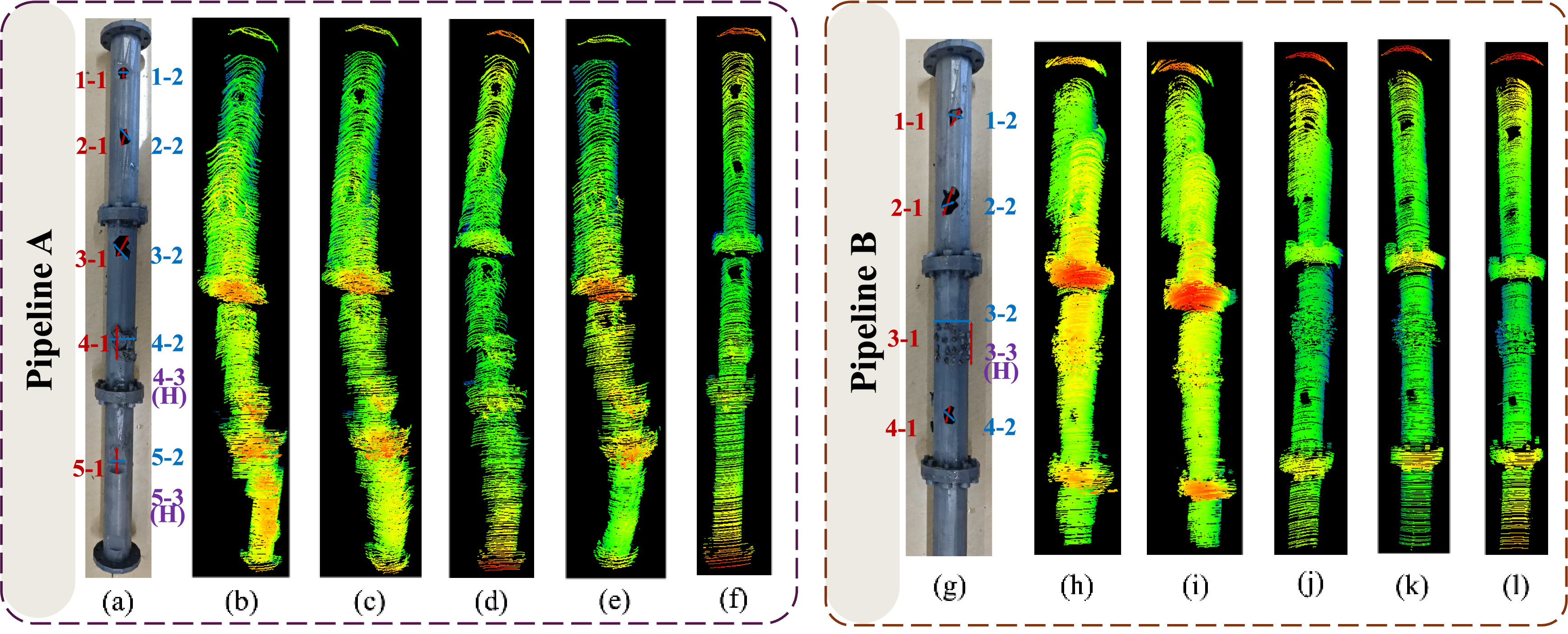}
	
	\caption{Visualization of imaging results by different methods. (a) and (g) Original image. (b)-(f) and (h)-(l) Imaging results of Pipeline A and Pipeline B from DR, AEKF, ICP, GICP and the proposed method.}
	\label{fig20}
\end{figure*}
\indent After fixing the threshold, the registration results of different methods are summarized in \autoref{table13} and \autoref{table14}. The \autoref{table13} indicates that the proposed method achieves an RMSE of 1.08 for Pipeline A, representing reductions of 86.2\% and 84.6\% compared with ICP and GICP, respectively. For Pipeline B, the RMSE is 2.31, which is 67.9\% and 63.0\% lower than those of ICP and GICP, respectively. The overall registration results in the \autoref{table14} further demonstrate that the proposed method outperforms the other methods. It is worth noting that in global registration, the RMSE of ICP and GICP sometimes exceed those of DR and AEKF. This occurs because partial overlaps remain between consecutive pipeline point clouds after multi-frame alignment, leading to local interpolation that slightly reduces numerical precision. However, the primary purpose of registration is not only to minimize the RMSE but also to maintain the geometric integrity of the pipeline under variable-speed motion. Although the numerical error caused by overlap may increase slightly, the proposed registration method significantly improves shape consistency and the visibility of local defects. It enable more stable and realistic point cloud under dynamic conditions. As illustrated in \autoref{fig20}, the proposed method better restores the original geometric appearance of the pipeline. 
\subsection{Pipeline defect detection experiments}
\indent To further validate the effectiveness of the proposed ED-ICP algorithm in actual scenes, the defect detection experiments are performed on multi-defect pipelines A and B. As shown in \autoref{fig20}, the DR and AEKF methods suffer from cumulative drift and non-uniform motion, leading to noticeable geometric deformation of the point cloud. Although the ICP and GICP methods enhance inter-frame registration consistency, local distortion and surface blurring remain at the junctions of adjacent imaging frames. It will reduce the imaging accuracy of small defect regions. In contrast, the proposed method preserves the geometric continuity of the entire pipeline, providing more complete and accurate 3D representations of local defect morphologies. The generated point cloud exhibits a compact shape with clear boundaries and shows strong visual consistency with the actual pipeline structure. The \autoref{table15} presents the quantitative detection results.  For pipelines A and B, the most significant deviations occur in the defect height of A4-3 (attachment), A5-3 (depression), and B3-3 (attachment). These deviations primarily arise from significant variations in local surface normals within small-scale defect regions. These variations modify the local reflection direction, thereby inducing geometric discrepancies between the incident and imaging rays. In addition, the stripe edges may become blurred due to underwater light scattering. The precision of light plane estimation is thereby degraded, further amplifying the height measurement error. Given the small actual height of the defects, even minor measurement noise or systematic errors can produce large deviations in the measured height values. The relative error of defects is below 4.10\% for Pipeline A and 5.67\% for Pipeline B when the height value is excluded. These results demonstrate that the proposed method effectively suppresses cumulative registration deviations and local surface deformations, accurately restoring the geometric dimensions of defect regions.
\begin{table*}[t]
	\footnotesize
	\centering
	\caption{ Experimental results of pipeline defect detection.}
	\resizebox{\textwidth}{!}{
		\begin{tabular}{cccccc}
			\toprule[1.5pt]
			Class &Number &
			Groud truth/mm & Measurement value/mm&  Absolute error/mm &  Relative error/\%\\ 
			\midrule    
			Pipeline A & $1-1$ & 12 &11.94&0.06 &0.50 \\
			$ $ & $1-2$ & 11 &11.05&0.05 &0.45\\
			$ $ & $2-1$ & 16 &16.06 &0.06 &0.38\\
			$ $ & $2-2$ & 17 & 17.13 &0.13& 0.76 \\
			$ $ & $3-1$ & 23 &23.21 &0.21& 0.91 \\
			$ $ & $3-2$ & 21 &21.08 &0.08 & 0.38 \\
			$ $ & $4-1$ & 30 &29.51 &0.49 & 1.63\\
			$ $ & $4-2$ & 35 & 35.63 &0.63 &1.80 \\
			$ $ & $4-3$ & 5 &5.59 &0.59 & 11.80\\
			$ $ & $5-1$ & 27 &26.12 &0.88 &3.26 \\
			$ $ & $5-2$ & 21 &20.14 &0.86 & 4.10\\
			$ $ & $5-3$ & 3.5 &3.98 &0.48 &13.71 \\
			
			Pipeline B & $1-1$ & 15 &15.85 &0.85 &5.67\\
			$ $ & $1-2$ & 13&13.03 &0.03 &0.23 \\
			$ $ & $2-1$ & 21 & 21.06 &0.06 &0.29\\
			$ $ & $2-2$ & 12 &12.48 &0.48 &4.00 \\
			$ $ & $3-1$ & 33 &33.98 &0.98 &2.97 \\
			$ $ & $3-2$ & 31 &31.88 &0.88 &2.84\\
			$ $ & $3-3$ & 5 &4.26 &0.74 &14.80 \\
			$ $ & $4-1$ & 15 &14.42 &0.58 &3.87\\
			$ $ & $4-2$ & 12 &11.69 & 0.31 & 2.58\\
			
			\bottomrule[1.5pt]	
		\end{tabular}%
	}
	\label{table15}%
\end{table*}%
\section{Conclusions and prospect}
\label{section8}
\indent In this paper, the UW-SLD system was developed, integrating multi-mode structured light imaging, multi-source information fusion, and intelligent point cloud registration. The system achieved key technological breakthroughs in underwater 3D imaging, including distortion correction, heterogeneous information fusion, intelligent detection and multi-scenario applications. Firstly, a camera calibration method based on the refraction model was established. The distortion correction method greatly improved the computational efficiency while maintaining high precision. Secondly, the extrinsic parameter calibration method combining structured light scanner and DVL significantly improved the coordinate transformation accuracy. Subsequently, a hierarchical multi-frequency information fusion strategy was proposed for heterogeneous information fusion. In terms of pose estimation, the AEKF method was introduced to fuse inertial and acoustic information, enabling stable output even in the case of DVL signal interruption. Based on this, a multi-mode structured light imaging strategy including translation, rotation, and translation-rotation was designed to cope with different detection scenarios. The experimental results show that the relative error of the method in each imaging mode is less than 3\%. Although the translation-rotation mode is complex, its accuracy does not significantly decrease. This verified the stability and adaptability of the method. To achieve high-precision imaging in dynamic scenarios, this paper proposed the ED-ICP algorithm for point cloud registration. The algorithm effectively suppresses noise while preserving the structural features of the pipeline, achieving an optimal balance between accuracy and integrity. Finally, the capability of the system to accurately detect the fine defect structures of the pipeline was validated through the defect detection experiment. In summary, this paper developed a imaging system capable of capturing both the complete macroscopic morphology and fine defect structures of underwater pipelines. Comprehensive experimental results demonstrated that the system exhibited strong reliability and considerable potential for application in high-precision underwater detection. It can provide critical technical support for the regular maintenance of underwater pipelines.\\
\indent In the future, we will continue to optimize and upgrade this system. In the final defect detection stage, we plan to further investigate defect classification methods based on deep learning. Moreover, a practical approach for calculating the defect area of pipelines will be explored to achieve fully automated system detection.Ultimately, the system will be integrated into various underwater robots to perform underwater pipeline detection tasks across multiple scenarios.
\section*{CRediT authorship contribution statement}
\indent \textbf{Qinghan Hu:} Conceptualization, Methodology, Software, Formal analysis, Visualization, Writing–original draft. \textbf{Haijiang Zhu:} Conceptualization, Methodology, Investigation, Resources, Writing—review and editing, Supervision. \textbf{Na Sun:} Methodology, Investigation, Resources, Writing—review and editing. \textbf{Lei Chen:} Software, Formal analysis, Validation. \textbf{Zhengqiang Fan:} Validation, Writing—review and editing. \textbf{Zhiqing Li:} Resources, Supervision, Project administration,  Funding acquisition.
\section*{Declaration of competing interest}
\indent The authors declare that they have no known competing financial interests or personal relationships that could have appeared to influence the work reported in this paper. 
\section*{Acknowledgement}
This work was supported by National Key Research and Development Program of China (2024YFB4710701).
\section*{Data availability}
\indent Data will be made available on request.

}\end{multicols}
\end{document}